%% file: arxiv.tex
\documentclass[11pt]{article}

\usepackage[preprint]{acl}

\usepackage{times}
\usepackage{latexsym}

\usepackage[T1]{fontenc}

\usepackage[utf8]{inputenc}

\usepackage{microtype}

\usepackage{inconsolata}

\usepackage{graphicx}
\usepackage[utf8]{inputenc} 
\usepackage[T1]{fontenc}    
\usepackage{hyperref}       
\usepackage{url}            
\usepackage{booktabs}       
\usepackage{colortbl}        
\usepackage{amsfonts}       
\usepackage{nicefrac}       
\usepackage{microtype}      
\usepackage{xcolor}         
\usepackage{amsmath,amssymb}
\usepackage{wasysym}
\usepackage{graphicx}
\usepackage{booktabs}
\usepackage{multirow}
\usepackage{longtable}
\usepackage{array}
\usepackage{ragged2e}
\usepackage{url}
\usepackage{xcolor}
\usepackage{titlesec}
\usepackage{etoc}
\usepackage{algorithm}
\usepackage{algpseudocode}
\usepackage{amssymb}   
\usepackage{tabularx}  
\usepackage{booktabs}  
\usepackage{adjustbox} 
\usepackage{mdframed}
\usepackage{float}
\usepackage{needspace}
\usepackage{enumitem}

\newenvironment{promptframe}{%
  \Needspace{10\baselineskip}%
  \begin{mdframed}[
  backgroundcolor=gray!10,
  linecolor=black!40,
  linewidth=0.8pt,
  roundcorner=5pt,
  skipabove=5pt,
  skipbelow=6pt,
  innertopmargin=5pt,
  innerbottommargin=5pt,
  innerleftmargin=6pt,
  innerrightmargin=6pt,
  splittopskip=0.6\baselineskip,
  splitbottomskip=0.4\baselineskip
  ]%
  \small
  \setlength{\parindent}{0pt}%
  \setlength{\parskip}{2pt plus 1pt minus 1pt}%
}{%
  \end{mdframed}%
}
\newenvironment{promptitemize}{%
  \begin{itemize}[leftmargin=1.2em,itemsep=1pt,topsep=2pt,parsep=0pt,partopsep=0pt]%
}{%
  \end{itemize}%
}

\title{LongChart VQA: A Comprehensive Benchmark for MLLMs with \\ Complex Multi-Chart Reasoning}

\author{Ziyan Xiao, Yinghao Zhu, Wenting Zhang, Heaju Kim, Lequan Yu\thanks{Correspondence to Lequan Yu (\texttt{lqyu@hku.hk})} \\
  The University of Hong Kong}

\begin{document}
\maketitle

\begin{abstract}
Multimodal large language models (MLLMs) rapidly evolve with expanding context and strengthened reasoning to support multi-chart understanding and multi-step inference. These abilities become increasingly important as MLLMs are adopted in complex agentic tasks. However, existing benchmarks largely emphasize single-chart perception, and simple chart-to-chart connections remain insufficient for evaluating MLLMs  might be insufficient in measuring MLLM's capability in these aspects. 
To capture multi-chart complexity while ensuring consistency and validity, we design a latent-graph–supported synthesis pipeline. Building on this pipeline, we introduce \textbf{LongChart}, a benchmark that extends VQA sets to an average of 6.5 images and 31.2 questions. We evaluate 10 SOTA MLLMs on this benchmark, and further examine three aspects influencing performance, including reasoning patterns, auxiliary tools, and robustness against image perturbations. 
Our results show that MLLM accuracy decreases and varies significantly as computational complexity increases, which offer a direction for future research in multi-chart reasoning.
\end{abstract}

\section{Introduction}\label{sec:introduction}
Multimodal large language models (MLLMs) have demonstrated strong reasoning ability on a wide range of tasks, including commonsense understanding~\cite{yue2024mmmu,lu2022learn,marino2019ok,yue2025mmmu}, mathematical reasoning~\cite{lu2023mathvista, zhang2024mathverse, wang2024measuring}, and visual understanding across diverse image sources. Driven by advances such as multimodal chain-of-thought prompting~\cite{zhang2023multimodal} and reinforcement learning~\cite{guo2025deepseek,team2025kimi}, MLLMs are moving beyond single-image interpretation toward more complex cross-image and long-context settings~\cite{fu2023mme, kil2024mllm,huang2026vision}. In these scenarios, maintaining precision in understanding across multiple images with intricate relationships among them become central challenges.


\begin{figure}[!t]
\centering
\includegraphics[width=\columnwidth,trim=255 0 130 10,clip]{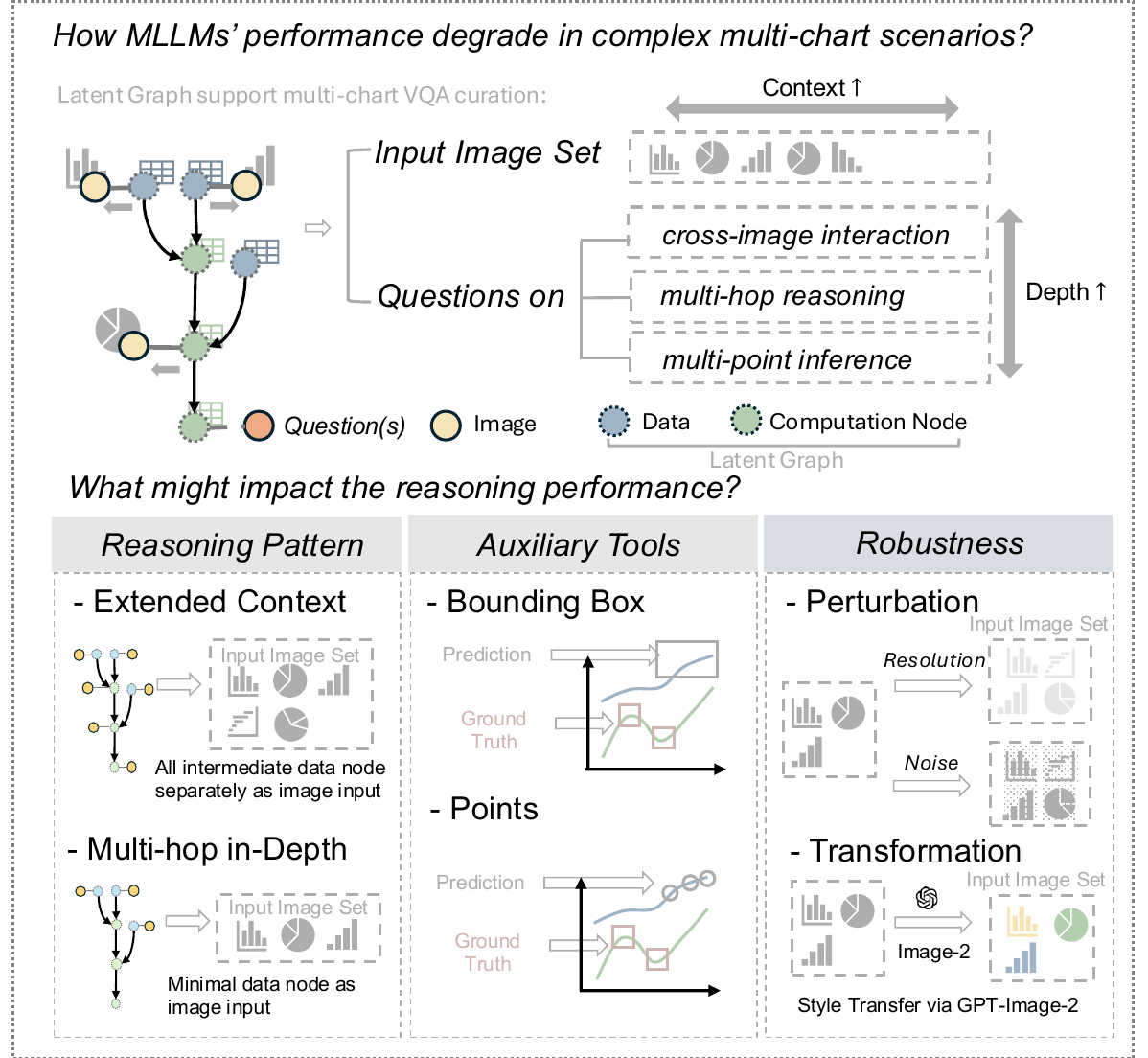}
\caption{Motivation of this work}
\label{fig:method_overview}
\end{figure}

\input{tables/literature_comparison}

Multi-chart, as a common medium for rich information, remains an underexplored area for MLLM's performance evaluation. Pioneering works of chart-related benchmarks \cite{masry2022chartqa, methani2020plotqa}, primarily evaluate basic chart understanding, such as element identification and data retrieval. The follow-up works focuses on enhancing task complexity by complicated scientific plots~\cite{wang2024charxiv,foroutan2025wikimixqa,shen2024sci_cqa}, injecting misleading information\cite{chen2025unmasking}, or entend to multilingual settings\cite{xu2025polychartqa}.

Some recent works have begun to explore multi-chart relationships, such as MultiChartQA~\cite{zhu2025multichartqa}, which focuses on extraction and comparison across charts, and InterChart~\cite{iyengar2025interchart}, which increases task difficulty by synthesizing additional charts from single-chart datasets. However, these approaches often building up complexity by LLM-based question generation and open-ended answers, which can introduce ambiguity in evaluation. More importantly, the relationships between multiple charts that distinguishes multi-chart QA from single-chart tasks, especially considering professional multiple chart analysis involves intensive data computation and analysis. The automated synthesizing or grouping in previous work may leave the benchmark lack of evaluation on these skills. To address these challenges, we design a latent graph-supported curation pipeline that provides a structured foundation for constructing complex multi-chart benchmarks.

One of the central challenges is ensuring that charts are grounded in a shared and internally consistent data space, so that cross-image interactions yield meaningful results. To ensure the internal consistency, the synthetic pipeline is designed with the following properties: (1) \textbf{latent graph as backbone}: data and images are defined and connected with logical relationship; (2) \textbf{data consistency}: data is synthesized only with a minimal set of nodes, with propagation to the rest to ensure uniqueness; and (3) \textbf{difficulty through a growing hops not answer length}: through close-end question, but the difficulty of questions is distributed with the number of hops, avoiding reliance on external knowledge or open-ended judgments.

Based on the above synthesizing framework, we propose \textsc{LongChart} benchmark, a multiple chart VQA benchmark for evaluating the reasoning capabilities of MLLMs. The benchmark includes 557 images and 2,876 questions. We further largely extending the number of question for each set of images to study change of performance with question complexity. In the main set, each VQA instance contains on average 6.5 images, 31.2 questions, and approximately 716.8 data points. We evaluate 10 state-of-the-art MLLMs across these questions and further analyze three aspects that influence model reasoning, reasoning patterns, the use of auxiliary tools such as grounding and pointing, and robustness under image perturbations (Figure~\ref{fig:method_overview}). Results show that when computational or analytical complexity increases, MLLM accuracy declines sharply, which may highlighting directions for future exploration.

In summary, we aim to realize the following objectives:
\begin{itemize}
    \item We propose a latent graph–supported pipeline for synthesizing multi-chart, multi-question datasets, that overcomes the challenges in the benchmarking of complex computation ability in multi-chart problems.
    \item We construct LongChart Benchmark and evaluate 10 MLLMs reasoning ability when largely extends the context and reasoning depth of multi-chart question answering.
    \item We provide a comprehensive analysis of model performance from three aspects, the impact of reasoning patterns, the role of auxiliary tools such as grounding and pointing, and robustness against common image perturbations and transformations.
\end{itemize}


\begin{figure*}[!ht]
\centering
\includegraphics[width=\textwidth,trim=30 15 10 40,clip]{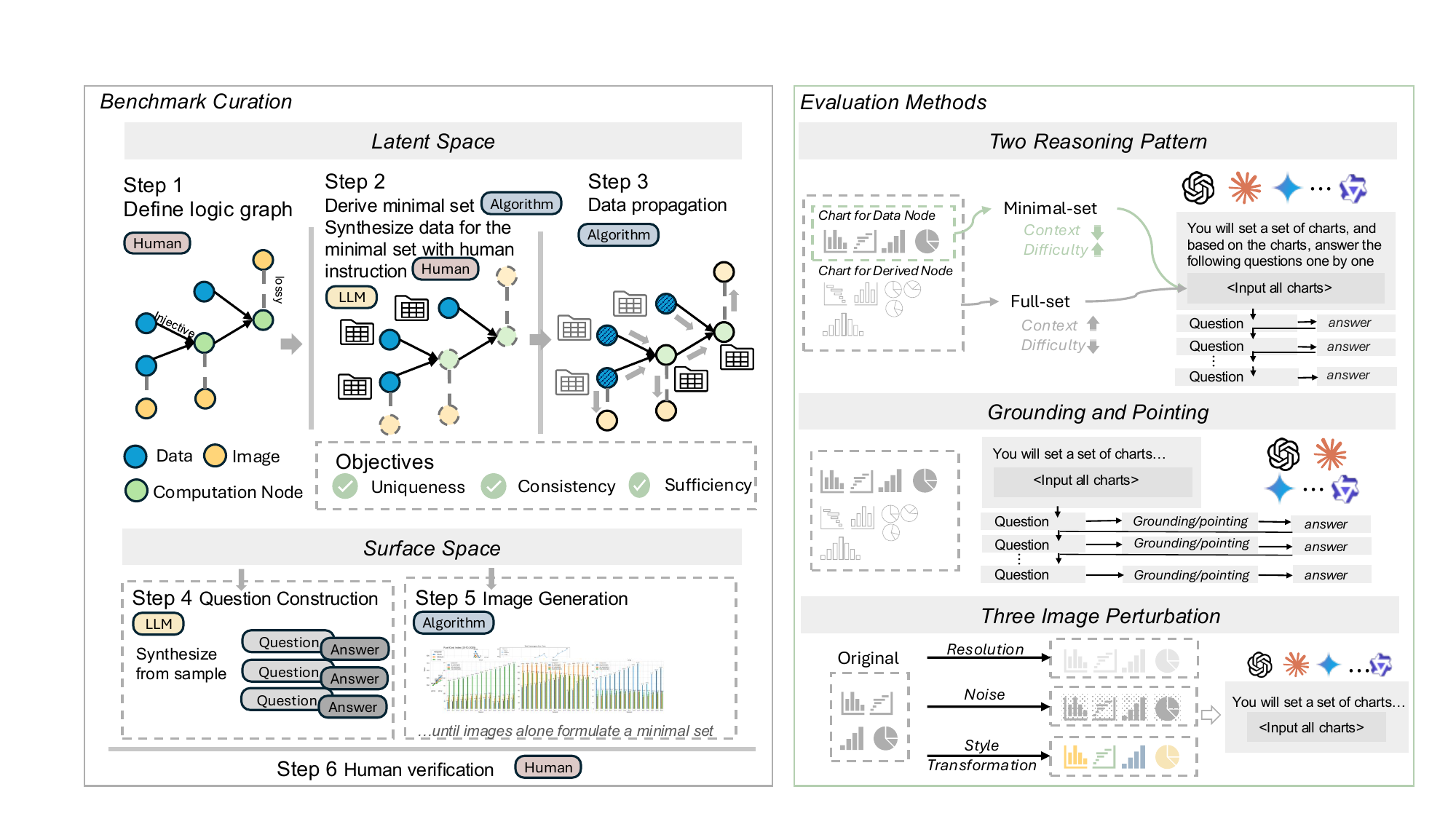}
\caption{\textbf{Overview of the VQA set curation process and evaluation.} The pipeline comprises graph design, data synthesis, and VQA generation with human-in-the-loop quality control.}
\label{fig:vqa_curation}
\end{figure*}

\section{Related Work}\label{sec:related_work}
\subsection{Chart-Related Benchmarks}\label{subsec:chart_benchmarks}
Chart is a prevalent area of focus before the prevalence of MLLMs. Early chart-related benchmarks such as ChartQA~\cite{masry2022chartqa} and PlotQA~\cite{methani2020plotqa} established the foundation of this area. Subsequent benchmarks, including ChartBench~\cite{xu2023chartbench}, ChartXiv~\cite{wang2024charxiv}, ChartX~\cite{xia2025chartx}, and Sci-CQA~\cite{shen2024sci_cqa}, further extended evaluation to more complex figure formats and reasoning skills. With the MLLM's reasoning capabilities radidly improves,
more recent benchmarks have started to explore extended and intergrated capabilities of chart reasoning, including multilingual settings~\cite{xu2025polychartqa}, multiple-chart reasoning~\cite{zhu2025multichartqa}, and the visual insights~\cite{tang2025chartmuseum}. Multi-image benchmarks has become prevalent in other domains. Existing works in this area includes, in-depth reasoning questions~\cite{fan2025depth}. Despite the progress, two main challenges in multi-chart VQA benchmarks stayed insufficiently explored. First, current benchmarks may yet to push sufficiently toward the long-context settings so that there saw significantly change in the reasoning pattern from pure details, to a global-and-detail thinking pattern. Second, for such a complex Multi-Chart benchmark, it requires more holistic evaluation than one simple question to accurately measure the boundary of reasoning performance, and indicate the factors that impacts the reasoning capabilities.

\subsection{MLLM Reasoning with Chart Understanding}\label{subsec:mllm_reasoning}
Chart understanding is an essential capacity to measure MLLM's reasoning capability. Early works in improving MLLM's Chart understanding focused on curating large scale chart images, and post-training the MLLM models, through instruction tuning\cite{han2023chartllama, zeng2024advancing}, and supervised fine-tuning\cite{meng2024chartassistant}. Moreover, emergent works also focused on breaking the modality constraints of charts, such as Matcha\cite{liu2023matcha} propose chart-to-table alignment, ChartMoE\cite{xu2024chartmoe} propose alignment between chart-table-JSON-code.
There are increasing attempts in MLLM reasoning aimed to improve chart understanding via enhanced tools or fine-grained capabilities, including accurate parsing\cite{li2026visual, xingchendavinci}, grounding\cite{xu2025chartpoint}, and automatic edition or generation\cite{zhao2025chartedit}. However, most of the works focused on precisely local unit of information, such as data extraction due to their prevalency in early benchmarks. It remains unclear that when moving to multi-chart scenarios, how might MLLMs performance retain or decrease when changing from simple precision-oriented reasoning, to complex depth-oriented reasoning.

\begin{figure*}[!t]
\centering
\includegraphics[width=\textwidth,trim=125 10 75 80,clip]{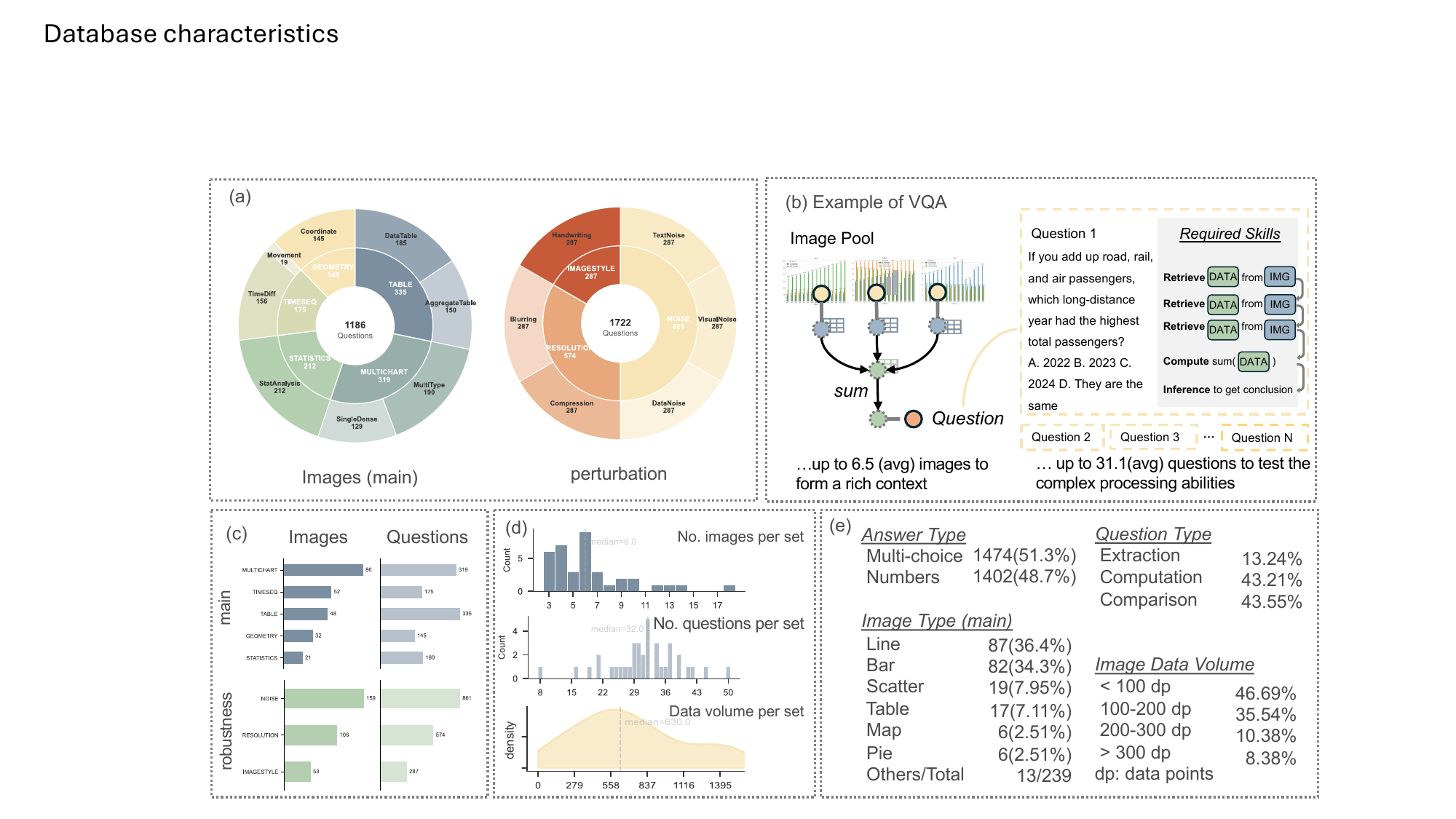}
\caption{\textbf{Dataset characteristics of LongChart VQA.} (a) Question distribution across curation categories. (b) A sample question; (c) Image and question count. (d) Distribution of number of images, questions and data points per QA set; (e) Statistics in the VQA.}
\label{fig:dataset_characteristics}
\end{figure*}

\section{Benchmark Curation}\label{sec:benchmark_curation}
This benchmark provides comphensive evaluation on MLLMs' understanding in multi-chart complex reasoning, and investigate the performance under reasoning patterns, auxilary tools and robustness testing. To enable the extension of context and reasoning difficulty, as well as ensure the consistency of answer, we designed the following latent-graph driven VQA set curation process. In the following, we present the design for VQA dataset curation ~\ref{subsec:vqa_curation}, validation ~\ref{subsec:human_in_the_loop}, and dataset characteristics ~\ref{subsec:benchmark_characteristics}. The full VQA curation process is provided in Appendix~\ref{sec:appendix_curation_theory}. Details of human-in-the-loop curation plat form is in Appendix~\ref{sec:appendix_annotation}. Some samples of VQA are provided in Appendix~\ref{sec:appendix_examples}

\subsection{Latent Space That Supports VQA Curation}
\label{subsec:vqa_curation}
\paragraph{Graph-as-backbone.}
To ensure consistency and meaningfulness across charts, we design the graph as the backbone of the VQA curation process. As shown in Figure~\ref{fig:vqa_curation}, the graph is manually constructed to abstract real-world data relationships—for example, between GDP, inflation, and price index. This serves as a primary step in ensuring coherence across multiple data sources and is later used for data synthesis and propagation.

Formally, for each question set we define a directed acyclic graph (DAG)
\[
G = (V, E),
\]
where $V$ denotes the union of data nodes, computation nodes (intermediate variables derived from data), and image nodes (visual representations of other nodes).

Edges $E$ encode information flow and transformations between vertices. In practice, deterministic operators transform data nodes into computation nodes, capturing common relationships such as summation, difference, ratio, proportion to totals, time-shift, or spatial distance.

Information preservation is a critical attribute attached to each edge. It governs graph traversal for identifying minimal sets in data propagation and is defined in two categories:

\begin{itemize}
    \item An edge is \emph{bidirected}, denoted $u \leftrightarrow v$, if the transformation is reversible between $u$ and $v$. For example, a bar chart fully representing the values and distribution of a data node.
    \item An edge is \emph{directed}, denoted $u \rightarrow v$, if the transformation is one-way and $u$ cannot be reconstructed from $v$. For instance, ratios between data nodes or year-on-year changes relative to the original data.
\end{itemize}

\input{tables/question_type_accuracy_new.tex}

\paragraph{Minimal set and data propagation.}
Unlike simple chart curation, multi-chart data must satisfy two constraints: (1) \textbf{Uniqueness and Consistency}. Information across charts must remain coherent and free of conflicts; (2) \textbf{Sufficiency}. The information revealed must be adequate to answer all curated questions. The latent graph design supports these requirements as follows.

The information-preservation property of edges ensures that graph traversal yields a minimal set containing all information conveyed across charts, questions, and data in latent space. Nodes in this minimal set are unique and independent; thus, only these nodes are synthesized, as shown in Figure~\ref{fig:vqa_curation}(step 2). Data synthesis is performed via LLM calls guided by human-defined trends to strengthen practical relevance, while remaining nodes are deterministically computed through predefined transformations. This design guarantees uniqueness and consistency while reducing computational cost.

Sufficiency is enforced by assessing whether charts alone form a minimal generative set of information. Once data curation is finalized, images and QA pairs are generated independently, with edges directed or bidirected to latent data nodes depending on preservation properties. If gaps remain, the algorithm automatically compensates until charts alone constitute a sufficient minimal set, after which the curated VQA set undergoes human validation.



\subsection{Human-in-the-Loop Benchmark Construction}
\label{subsec:human_in_the_loop}
Although the latent-space design and algorithmic procedures ensure the validity of the complex VQA set, a key challenge for synthesized charts is the concern that they may appear disconnected from practical use. To address this, we incorporate a human-in-the-loop process to enhance meaningfulness, usefulness, and aesthetics.

\paragraph{VQA curation.} As shown in Figure~\ref{fig:vqa_curation}, human experts contribute at three critical stages of curation:
(1) defining the latent graph structure to ensure that connections between data nodes are realistic and grounded in real-world scenarios;
(2) specifying and refining aesthetic parameters during chart generation—such as image type, subplot layout, and color schemes—to align the synthetic charts more closely with practical usage;
(3) verifying and reviewing each curated QA set to confirm that questions are readable and answerable.

\paragraph{VQA validation.} After curation, external evaluators are invited to answer the questions using only basic tools such as a calculator. Their performance is compared against the ground truth to establish a human-level accuracy baseline.

\subsection{Benchmark Characteristics}
\label{subsec:benchmark_characteristics}
Figure~\ref{fig:vqa_curation} summarizes the question and image characteristics of the benchmark. The dataset contains 557 images and 2,876 questions across the main and robustness sets. Its long-context nature arises primarily from multiple charts and multiple questions within each VQA set. In the main set, each VQA instance includes on average 6.5 images (range: 3–18), 31.2 questions (range: 8–50), and 716.8 data points (range: 116–1,536). Hops are defined as the number of computation steps required to reach a conclusion. The average hop count is 1.41 (range: 0–8) in the main set and 1.35 (range: 0–8) in the robustness set.


\begin{figure*}[t!]
\centering
\includegraphics[width=\textwidth]{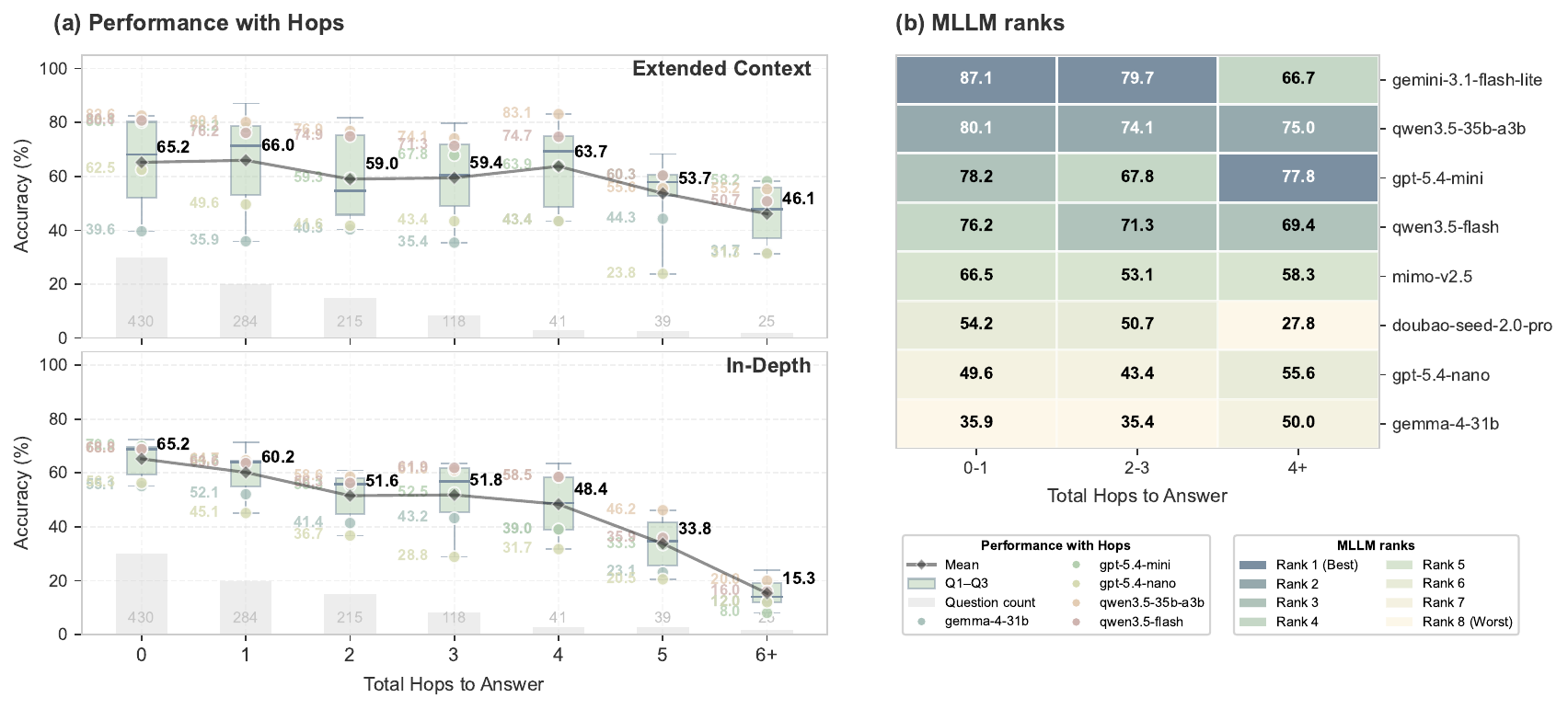}\\[6pt]
\caption{\textbf{Model performance as a function of reasoning hops.} (a) Accuracy degradation across increasing reasoning complexity under Extended Context (top) and In-Depth (bottom).(b) Performance ranks across model under different difficulty level of questions}
\label{fig:performance_by_hop}
\end{figure*}

\section{Experiments}\label{sec:experiment}
\subsection{Baseline Models}\label{subsec:baseline_models}

We evaluate a diverse suite of state-of-the-art MLLMs, spanning both proprietary APIs and open-source state-of-the-art models.
\textbf{Proprietary models.} We include Doubao-Seed-2.0-Lite\cite{ByteDanceSeed}, Doubao-Seed-2.0-Pro\cite{ByteDanceSeed}, Gemini-3.1-Flash-Lite\cite{GeminiFlashLite}, GPT-5.4-Mini\cite{openaiIntroducingGPT54}, GPT-5.4-Nano\cite{openaiIntroducingGPT54}, and Qwen-3.5-Flash\cite{deepmindGeminiPro}.
\textbf{Open-source models.} We additionally test Gemma-4-26B-IT\cite{deepmindGemma}, Gemma-4-35B-A3B\cite{deepmindGemma}, and Qwen-3.5-35B-A3B\cite{qwen3.5}.

The baseline configuration is provided in Appendix~\ref{sec:baseline_config}. Evaluation metric is defined in Appendix~\ref{sec:evaluation_metrics}. Prompts used in the baseline evaluation are listed in Appendix~\ref{sec:prompts}. 

Beyond the baseline, we design experiments to examine three dimensions that affect MLLM performance: (i) reasoning patterns, (ii) grounding and pointing as auxiliary tools, and (iii) robustness why implying image perturbations and transformation that approximate real-world settings. As illustrated in Figure~\ref{fig:vqa_curation}, these dimensions are defined in detail as follows.

\subsection{Reasoning Patterns}\label{subsec:reasoning_patterns}

\paragraph{Extended context.}
Visualizing intermediate variables in charts provides a shortcut for reducing reasoning complexity. As illustrated in Figure~\ref{fig:vqa_curation}(b), the Extended Context mode supplies visualizations for every computation node in the MLLM queries. This approach can be viewed as a redistribution of reasoning effort, shifting from multi-step inference to increased context length.


\paragraph{In-depth reasoning.}
As shown in Figure~\ref{fig:vqa_curation}(b), MLLMs can also answer all questions when provided with only a minimal set of charts. As discussed in Section~\ref{sec:benchmark_curation}, the latent-space graph ensures that such minimal sets of images are obtainable within the VQA dataset, containing the necessary information to answer each question. A comparison between Extended Context and In-Depth Reasoning highlights the reasoning patterns preferred by MLLMs.

\subsection{Grounding and Pointing}
Bounding boxes and point annotations are strong indicators of an MLLM’s visual reasoning on charts and can serve as intermediate outputs in multi-image inference. Evaluating their accuracy provides essential evidence of reasoning quality and supports multi-step inference. Leveraging the synthetic nature of our VQA set, ground-truth annotations for bounding boxes and points are readily available. For evaluation, we select 548 questions with clear visual cues. In both modes, the MLLM is prompted to return not only the final answer but also the bounding box or point corresponding to the key information. We report accuracy results and conduct bounding-box and point analyses. The prompts are provided in Appendix~\ref{sec:prompts}.

\begin{figure*}[t!]
\centering
\includegraphics[width=\textwidth, trim=40 110 0 15,clip]{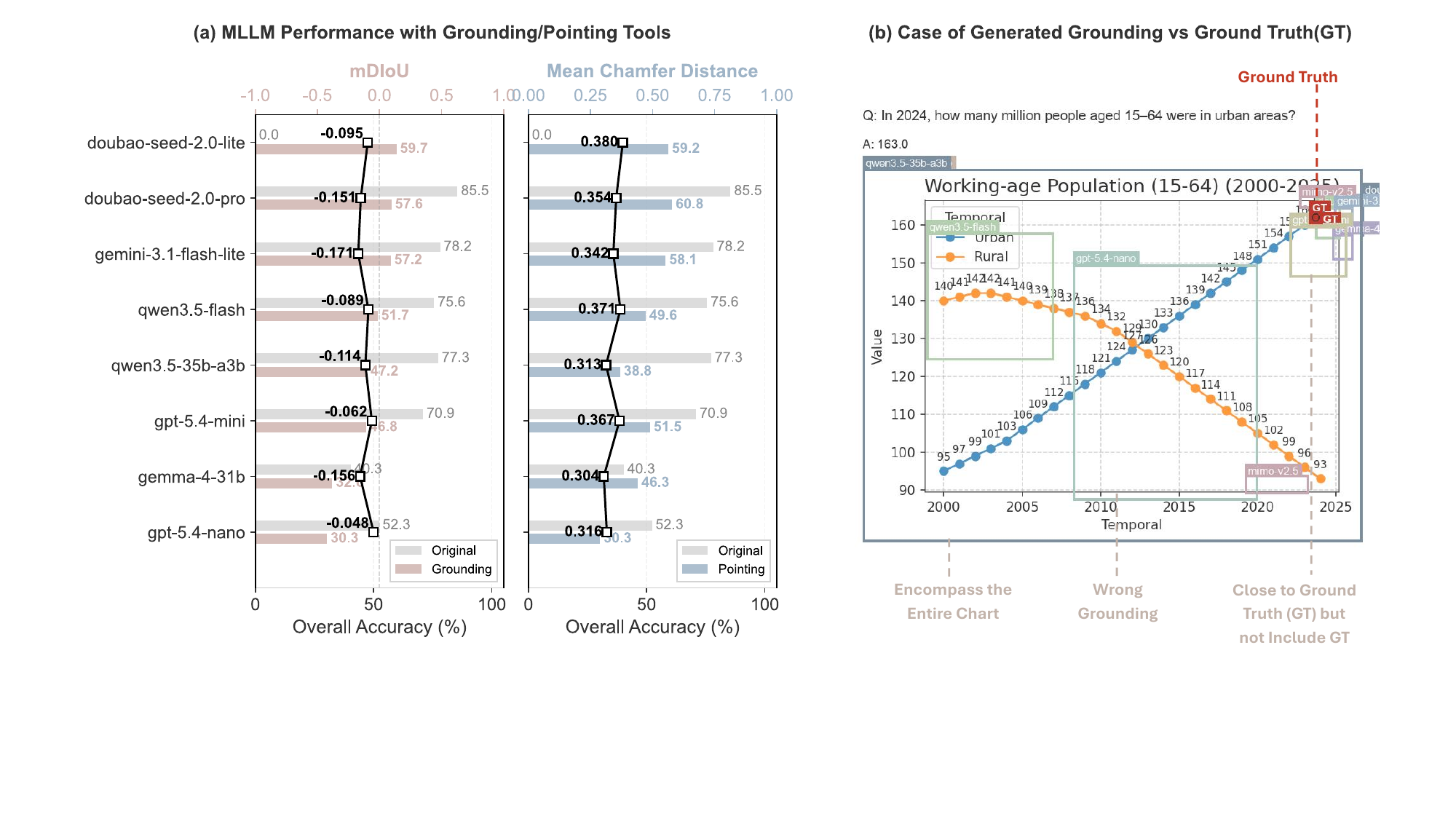}
\caption{\textbf{Performance with grounding and pointing.} (a) mDIoU and Chamfer distance indicate low quality in generated grounding box and point, with significant decrease in question answer accuracy. (b) a case study showing the prevalent patterns of low-quality grounding}
\label{fig:grounding_pointing}
\end{figure*}

\subsection{Robustness}\label{subsec:robustness}
To evaluate robustness, we select 10 VQA subsets comprising a total of 1722 questions and 318 images. These subsets are stratified across curated VQA categories, with average accuracy of the best-performing models ranging from 0.6 to 0.8. Each subset is subjected to seven types of image perturbations and transformations, with details in Appendix~\ref{sec:appendix_tasks}. The impact on MLLM performance is quantified by the accuracy difference between the original VQA set and its perturbed or transformed counterpart.

\section{Results}\label{sec:results}

\subsection{MLLM Performance}
Table~\ref{tab:question_type_accuracy} summarizes the performance of state-of-the-art MLLMs across different question types. Figure~\ref{fig:performance_by_hop} visualized the MLLMs performance distribution with total computational hops, and the rank of MLLM models under difficulty levels. To provide a holistic view, we also report MLLM performance across different VQA set types in Appendix~\ref{subsec:appendix_taxonomy}, and type of charts in Appendix~\ref{subsec:appendix_figure_types}.

\paragraph{Computation questions pose challenges.}
As shown in Table~\ref{tab:question_type_accuracy}, accuracy in multi-hop computation questions is consistently lower than overall performance. In fact, the gap between computation accuracy and that of extraction or comparison tasks show as a good indicator of a model’s reasoning ability across multiple images. Under in-depth input settings, the best-performing model, Doubao-seed-2.0 pro, showed a drop of $8.68\%$ compared to extraction tasks, while the gap widened to $35.75\%$ in GPT-5.4 nano and $30.52\%$ in Gemma-4-31b. The drop of accuracy in computation is a shared trend across the full set of image as input (Extended Context) and minimal set of image as input (In Depth), and is further verified in the following hop-related analysis.

\paragraph{Hops hinder performance more than context length.}
Table~\ref{tab:question_type_accuracy} further shows that models achieve higher accuracy in Extended Context settings than in In-Depth settings. Extended Context converts some multi-hop computations into intermediate visualizations, suggesting that reducing hops in exchange for longer context benefits performance. Figure~\ref{fig:performance_by_hop}(a) highlights this trend, as accuracy declines sharply once total hops grows. In Extended Context, average accuracy drops to $53.7\%$ at five hops and further falls to $46.1\%$ beyond six hops, whereas in In-Depth settings accuracy drops more significantly to $33.8\%$ at five hops and $15.3\%$ beyond six.

\paragraph{Data computation and conversion remain the task-level bottleneck.}
These difficulties manifest at the task level as persistent performance gaps. Tables~\ref{tab:appendix_taxonomy_view1} and \ref{tab:appendix_taxonomy_view2} show that MLLMs perform best on tasks with direct and explicit data, without complex inference. For example, table-based VQA tasks under Extended Context achieve the highest accuracies, reaching $93.43\%$ for the best-performing models. In contrast, tasks involving less direct information show significant drops, such as $82.99\%$ under In-Depth settings, or $68.57\%$ on time-sequence data with the same best-performing models.

\begin{figure}[!h]
\centering
\includegraphics[width=\columnwidth]{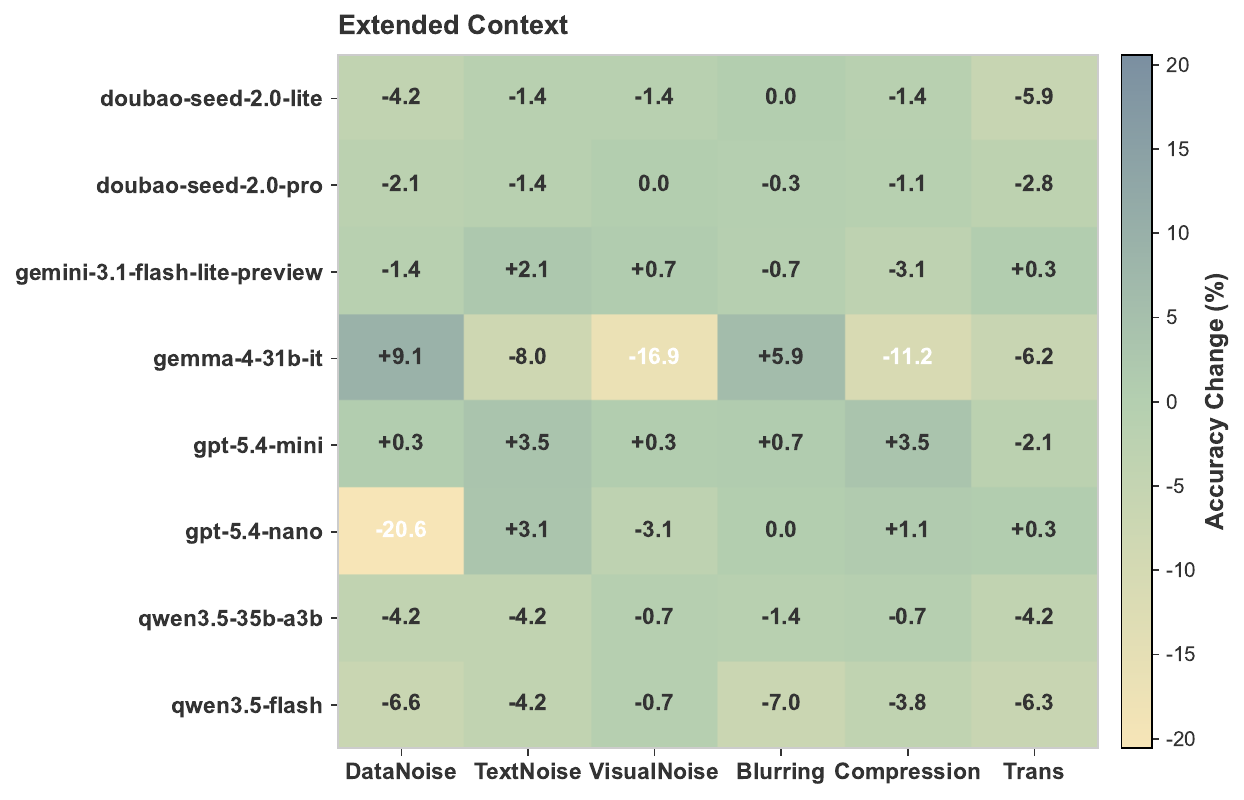}
\caption{\textbf{Impact of robustness perturbations.} The value shows equals the accuracy in the robustness set minus the accuracy for the corresponding original synthesized charts. Trans=Image Transformation.}
\label{fig:robustness_impact}
\end{figure}

\subsection{Analysis}

\paragraph{Grounding and pointing are inaccurate.}
Figure~\ref{fig:grounding_pointing}(a) illustrates both the low grounding quality and performance decline when MLLMs are required to generate bounding boxes or point annotations to cross-validate their answers. All models yield negative mDIoU scores, with mean Chamfer Distance ranging from $0.316$ to $0.380$. Case studies in Figure~\ref{fig:grounding_pointing} show that errors arise from outputs that are either completely irrelevant, approximate but imprecise, or overly broad such as circling the entire plot rather than the key area. Moreover, producing bounding-box or point coordinates often reduces overall performance significantly. For instance, generating grounding box causes best performing models drop from $85.5\%$ of accuracy to $57.6\%$. These findings highlight the difficulty of employing self-adjusted image tools as intermediate steps in agentic pipelines to improve multi-hop QA reasoning.


\paragraph{Accuracy deviation under robustness settings.}
Heterogeneous effects of image perturbation and transformation imposed across MLLMs. As shown in Figure~\ref{fig:robustness_impact}, robustness transformation impose not only negative effects, but also interestingly, some positive effects on model performance. Overall leading models present strong robustness against common image perturbations. Data noise and image transformation poses strong to model performance.

\begin{figure}[!h]
  \centering
    \includegraphics[width=0.95\linewidth]{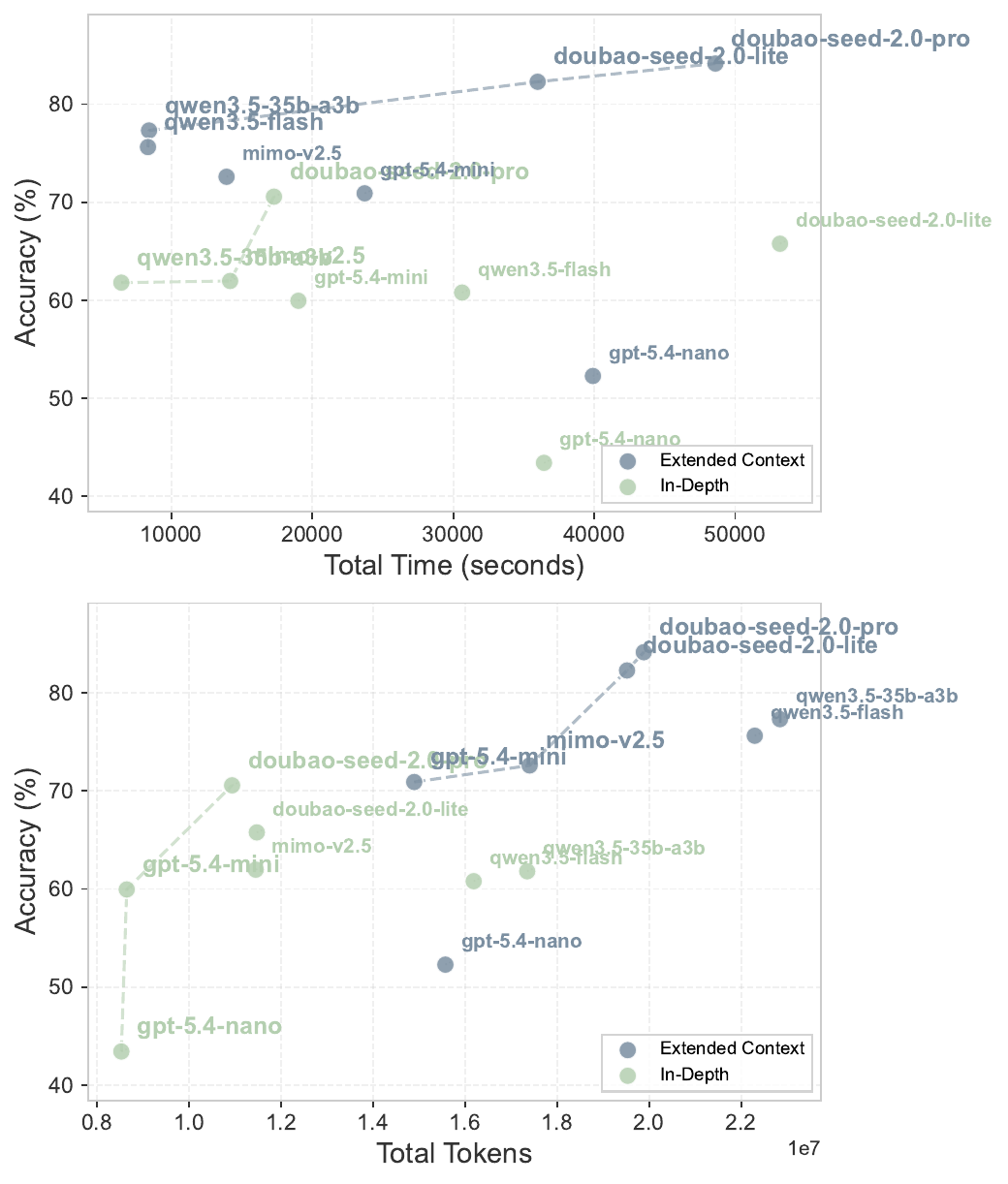}
  \caption{\textbf{Cost-performance distribution of evaluated models.} Two seperate efficient frontier curves of Extended Context and In-Depth presents, with both token count (top) and runtime (bottom) versus accuracy.}
  \label{fig:cost_performance}
  \end{figure}

\paragraph{Cost-performance frontier.}
Figure~\ref{fig:cost_performance} illustrates the cost–performance frontier of MLLMs under breadth and depth modes. 
Two clear efficient front-tier were shown both in the token measurement and time measurement, indicating space for improvement when extending the inference time and context. Switching from Extended Context mode to In-Depth mode, lead to 0.55x to 0.76x times reduction in tokens, and 0.36x to 3.68x change in procesing time. At present, the sacrifice of times and tokens in exchange for accuracy pattern is present.

\section{Conclusion}\label{sec:conclusion}
This paper introduces \textsc{Long-Chart Bench}, a new pipeline and benchmark designed to evaluate MLLMs’ visual reasoning and performance in multi-chart settings with complex computational relationships. Our evaluation of state-of-the-art MLLMs reveals a widening performance gap between MLLMs as data become less apparent due to multi-chart interaction and computational difficulty. We further observe that current models remain vulnerable when required to produce grounding outputs and when exposed to external perturbations. These findings underscore the need for future advances in MLLMs to strengthen visual reasoning in chart contexts, particularly as between-chart relationships grow more intricate.


\section*{Limitations}
We acknowledge the following limitations of this work and outline directions for future improvement.

\paragraph{Scope of the QA Benchmark.} Compared with other synthetic chart benchmarks, the number of images and questions in our dataset is relatively limited. This constraint arises from the human-in-the-loop design and the difficulty of validating each question set, which requires approximately 30 minutes per set (e.g., 5 images with 30 questions) and a total curation period of about 15 days. While the dataset size is modest, we hope to provide a preliminary yet holistic view of multi-chart scenarios with complex relationships, complementing with diverse comparative evaluations under difference baseline settings and image scenarios.

\paragraph{Coverage of Models.} Due to time constraints, the number of evaluated MLLMs is limited relative to the broad range of models currently available. We selected representative models from leading developers, but still the evaluation does not capture the full spectrum of existing SOTA models. We plan to continue expanding coverage to a wider set of models to provide a more comprehensive assessment.

\bibliography{refs}

\appendix

\input{appendix}

\end{document}

%% file: tables/literature_comparison.tex
\begin{table*}[ht]
\centering
\definecolor{sagegreen}{HTML}{B4CFB0}
\definecolor{paleyellow}{HTML}{F7E5B7}
\definecolor{mutedrose}{HTML}{CFB4B0}
\newcommand{\harveyfull}{\textcolor{sagegreen}{\CIRCLE}}
\newcommand{\harveyhalf}{\textcolor{paleyellow}{\LEFTcircle}}
\newcommand{\harveynone}{\textcolor{mutedrose}{\Circle}}
\caption{\textbf{Comparison of chart-based reasoning benchmarks.} Abbreviations: Extract = Data perception and extraction, Compare = Compare and analysis data pattern, Compute = Complex Data Computation, Ground. = Grounding, Point = Pointing, Eval. = Evaluation Format; \harveyfull = Full support, \harveyhalf = Partial support, \harveynone = No support.}
\renewcommand{\arraystretch}{0.1}
\small
\resizebox{\textwidth}{!}{%
\begin{tabularx}{\textwidth}{lcccccccc}
\toprule
\multirow{2}{*}{\textbf{Work}} & \multicolumn{3}{c}{CAPABILITIES} & \multicolumn{2}{c}{AUXILIARY TOOLS} & \multicolumn{2}{c}{VQA} \\
\cmidrule(lr){2-4} \cmidrule(lr){5-6} \cmidrule(lr){7-8}
& \textbf{Extract} & \textbf{Compare} & \textbf{Compute} & \textbf{Ground.} & \textbf{Point} & \textbf{Image} & \textbf{Eval.} \\
\midrule
ChartQA\cite{masry2022chartqa}        & \harveyfull & \harveynone & \harveyhalf & \harveynone & \harveynone & 1 & Finite Answer \\
ChartBench\cite{xu2023chartbench}     & \harveyfull & \harveynone & \harveyfull & \harveynone & \harveynone & 1 &  Finite Answer \\
ChartXiv\cite{wang2024charxiv}       & \harveyfull & \harveyfull & \harveyfull & \harveynone & \harveynone & 1--6+ &  Open Answer \\
MultiChartQA\cite{zhu2025multichartqa}   & \harveyfull & \harveyfull & \harveynone & \harveynone & \harveynone & 1-3 &  Open Answer\\
Ours           & \harveyfull & \harveyfull & \harveyfull & \harveyfull & \harveyfull & 3-18 &  Finite Answer \\
\bottomrule
\end{tabularx}%
}
\end{table*}

%% file: tables/question_type_accuracy_new.tex
\begin{table*}[!t]
\centering
\caption{\textbf{M2MChartBench question-type accuracy across Extended Context and In-Depth evaluation.} The best-performing model in each category is typeset in \textbf{boldface}, and the runner-up is indicated with \textit{underline}. Human baseline is assessed under full chart set (Entended Context).}
\label{tab:question_type_accuracy}
\setlength{\tabcolsep}{4pt}
\resizebox{0.95\textwidth}{!}{%
\begin{tabular}{lcccccccc}
\toprule
\multirow{2}{*}{\textbf{Model}} & \multicolumn{4}{c}{\textbf{Extended Context}} & \multicolumn{4}{c}{\textbf{In-Depth}} \\
\cmidrule(lr){2-5} \cmidrule(lr){6-9}
& \textbf{Extract} & \textbf{Compare} & \textbf{Compute} & \textbf{Total} & \textbf{Extract} & \textbf{Compare} & \textbf{Compute} & \textbf{Total} \\
\midrule
\rowcolor{gray!15}
\multicolumn{9}{c}{\textbf{Baselines}} \\
Human &  81.78\%  &  81.78\%  &  81.90\%  & 83.10\% & --- & --- & --- & --- \\
\midrule
\rowcolor{gray!15}
\multicolumn{9}{c}{\textbf{Proprietary MLLMs}} \\
doubao-seed-2.0-lite & \textit{\underline{84.37\%}} & 82.28\% & 81.14\% & \textit{\underline{83.32\%}} & 74.25\% & 72.28\% & 64.19\% & 69.20\% \\
Doubao-seed-2.0-pro & \textbf{86.25\%} & \textbf{84.57\%} & \textbf{86.57\%} & \textbf{85.76\%} & \textbf{84.49\%} & \textbf{84.03\%} & \textbf{77.16\%} & \textbf{81.22\%} \\
Gemini-3.1-flash-lite & 82.66\% & 75.38\% & 74.29\% & 79.37\% & 76.99\% & 75.29\% & 70.94\% & 73.74\% \\
GPT-5.4-mini & 69.65\% & 79.65\% & 68.57\% & 72.66\% & 75.93\% & 74.82\% & 56.53\% & 67.33\% \\
GPT-5.4-nano & 45.80\% & 63.89\% & 40.00\% & 50.79\% & 54.86\% & 55.92\% & 35.25\% & 47.11\% \\
Qwen3.5-flash & 74.33\% & 81.18\% & 71.14\% & 76.10\% & \textit{\underline{80.56\%}} & 76.57\% & 59.41\% & 69.98\% \\
\midrule
\rowcolor{gray!15}
\multicolumn{9}{c}{\textbf{Open-source MLLMs}} \\
Gemma-4-26b-a4b-it & 68.99\% & 69.31\% & 74.24\% & 69.50\% & 65.02\% &    59.09\% &     53.68\% &    58.03\% \\
Gemma-4-31b-it & 36.29\% & 42.96\% & 36.40\% & 38.36\% & 52.78\% & 45.43\% & 36.81\% & 42.92\% \\
Mimo-v2.5 & 76.67\% & 67.37\% & 65.52\% & 72.60\% & 76.62\% & 73.55\% & \textit{\underline{65.24\%}} & 70.53\% \\
Qwen3.5-35b-a3b & 77.68\% & \textit{\underline{82.82\%}} & 72.86\% & 78.71\% & 79.86\% & \textit{\underline{76.73\%}} & 64.17\% & \textit{\underline{71.94\%}} \\
\bottomrule
\end{tabular}%
}
\end{table*}

%% file: appendix.tex
\section{Extended Evaluation Results and Analysis}\label{sec:appendix_evaluation}

\subsection{Extended MLLM Performance Across Data Curation Taxonomy}\label{subsec:appendix_taxonomy}
This subsection extends the main benchmark comparison by reporting model performance across the detailed data curation taxonomy. We separate the results into complementary views so that readers can inspect model behavior on both the main and robustness splits in a more fine-grained manner.

\input{tables/appendix_taxonomy_view1.tex}

\input{tables/appendix_taxonomy_view2.tex}

\subsection{Performance Across Figure Types}\label{subsec:appendix_figure_types}
We further analyze performance variation across figure types. To provide an intuitive summary of category-level strengths and weaknesses, we visualize the results with radar plots for the two evaluation splits and reserve a dedicated table for a more detailed category-by-category comparison.

\begin{figure*}[tbp]
\centering
\begin{minipage}[t]{0.47\linewidth}
\centering
\includegraphics[width=0.95\linewidth]{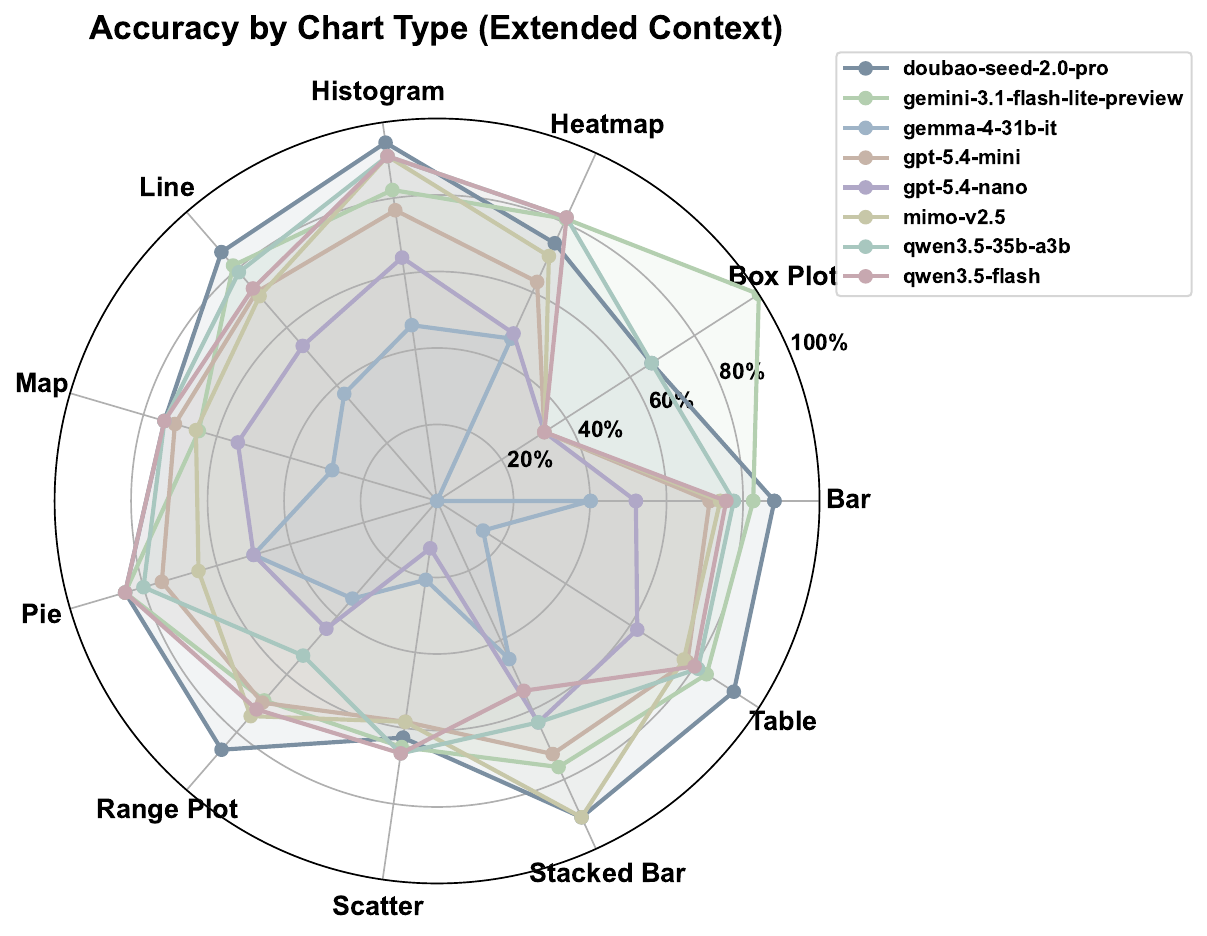}
\end{minipage}\hfill
\begin{minipage}[t]{0.47\linewidth}
\centering
\includegraphics[width=0.95\linewidth]{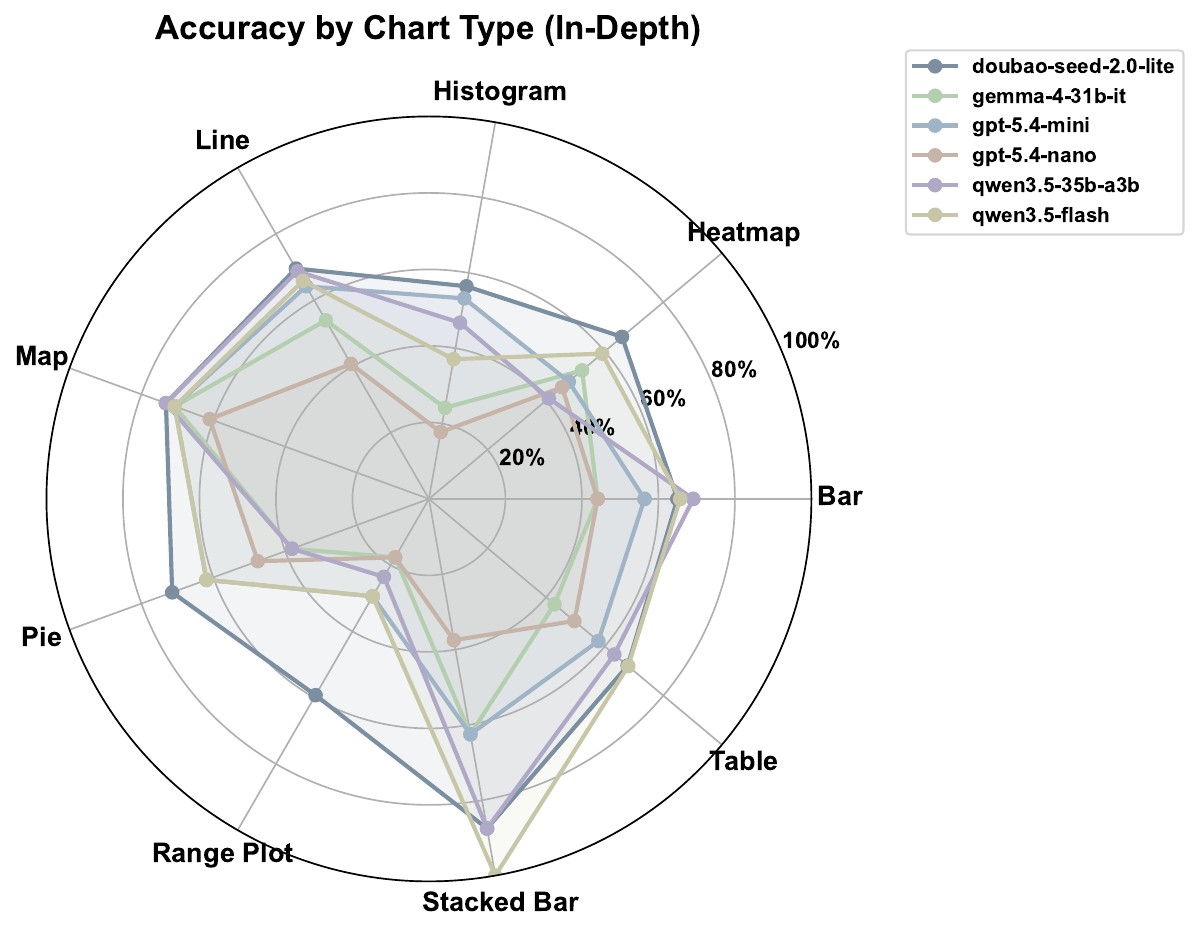}
\end{minipage}
\caption{Performance comparison across figure types on the two evaluation splits.}
\label{fig:appendix_figure_type_radar}
\end{figure*}

\input{tables/appendix_figure_type_breakdown.tex}

\subsection{Model Cost-Performance Distribution}\label{subsec:appendix_cost_perf}
Finally, we examine the trade-off between model performance and token cost on the main and robustness benchmarks. These scatter plots provide a compact view of which models achieve stronger accuracy under different cost budgets.


\begin{table*}[h]
\centering
\caption{\textbf{Accuracy change under visual perturbations.} Abbreviation: Trans = Transformation. (D) denotes In-Depth evaluation mode.}
\label{tab:robustness-perturbation}
\small
\begin{tabular}{lcccccc}
\toprule
\textbf{Model} & \textbf{Data Noise} & \textbf{Text Noise} & \textbf{Visual Noise} & \textbf{Blurring} & \textbf{Compression} & \textbf{Trans} \\
\midrule
doubao-seed-2.0-lite & $-$4.18\% & $-$1.39\% & $-$1.39\% & 0.00\% & $-$1.39\% & $-$5.92\% \\
doubao-seed-2.0-lite (D) & +24.59\% & +25.41\% & $-$3.70\% & +17.05\% & +15.91\% & +3.08\% \\
doubao-seed-2.0-pro & $-$2.09\% & $-$1.39\% & 0.00\% & $-$0.35\% & $-$1.05\% & $-$2.79\% \\
doubao-seed-2.0-pro (D) & +20.21\% & +19.16\% & +18.82\% & +19.16\% & +20.56\% & +17.42\% \\
gemini-3.1-flash-lite-preview & $-$1.39\% & +2.09\% & +0.70\% & $-$0.70\% & $-$3.14\% & +0.35\% \\
gemma-4-31b-it & +9.09\% & $-$8.02\% & $-$16.92\% & +5.88\% & $-$11.23\% & $-$6.17\% \\
gemma-4-31b-it (D) & $-$5.57\% & $-$4.53\% & $-$19.86\% & $-$0.70\% & $-$5.57\% & $-$7.32\% \\
gpt-5.4-mini & +0.35\% & +3.48\% & +0.35\% & +0.70\% & +3.48\% & $-$2.09\% \\
gpt-5.4-mini (D) & +12.20\% & +10.10\% & +7.67\% & +7.32\% & +14.63\% & +8.36\% \\
gpt-5.4-nano & $-$20.56\% & +3.14\% & $-$3.14\% & 0.00\% & +1.05\% & +0.35\% \\
gpt-5.4-nano (D) & $-$6.97\% & +5.57\% & 0.00\% & +8.01\% & +2.44\% & +2.44\% \\
mimo-v2.5 (D) & +11.50\% & +14.98\% & +12.54\% & +11.85\% & +11.50\% & +13.59\% \\
qwen3.5-35b-a3b & $-$4.18\% & $-$4.18\% & $-$0.70\% & $-$1.39\% & $-$0.70\% & $-$4.18\% \\
qwen3.5-35b-a3b (D) & +14.98\% & +9.06\% & +16.38\% & +6.97\% & +18.12\% & +12.89\% \\
qwen3.5-flash & $-$6.62\% & $-$4.18\% & $-$0.70\% & $-$6.97\% & $-$3.83\% & $-$6.27\% \\
qwen3.5-flash (D) & +16.72\% & +19.86\% & +20.56\% & +23.00\% & +18.82\% & +9.76\% \\
\bottomrule
\end{tabular}
\end{table*}

\subsection{Comparison of MLLM Answers Before and After Reasoning}\label{subsec:appendix_before_after}
This subsection compares model accuracy before and after explicit reasoning across different question types. The results help reveal whether performance gains mainly come from improved reasoning depth or from better answer grounding on specific question formats.

\begin{figure*}[tbp]
\centering
\includegraphics[width=0.85\linewidth]{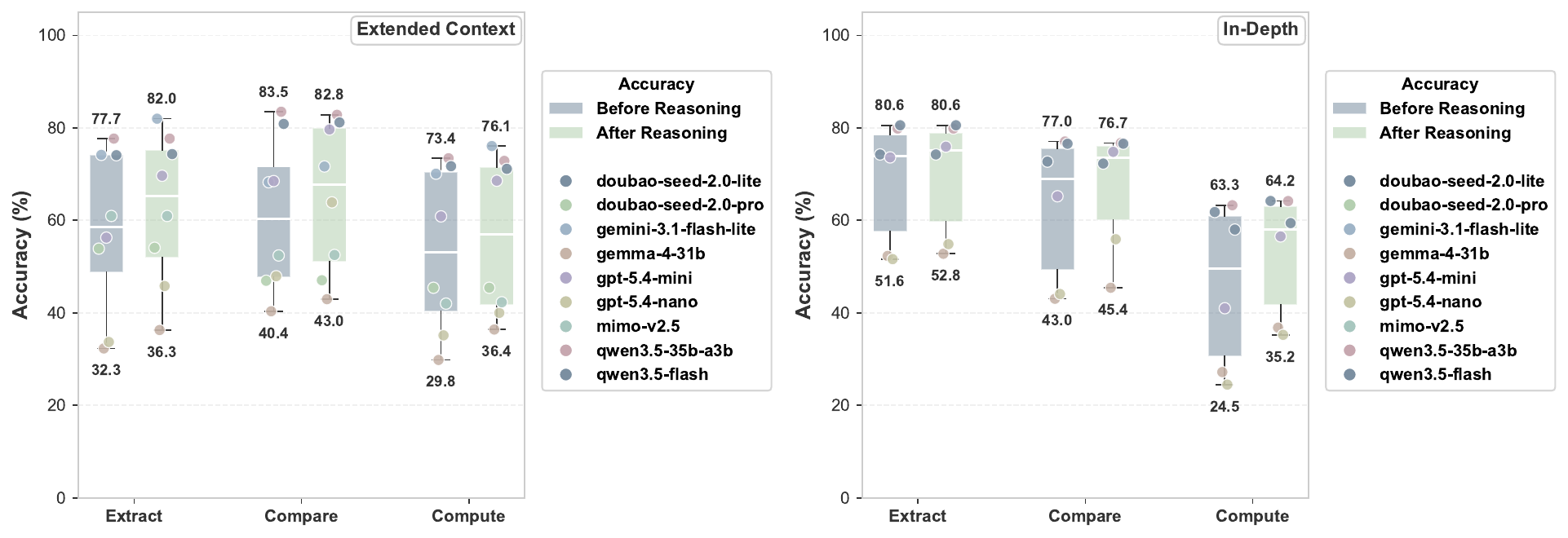}
\caption{Comparison of model accuracy before and after reasoning across question types.}
\label{fig:appendix_question_type_accuracy}
\end{figure*}

\input{tables/appendix_before_depth.tex}

\clearpage

\section{Task Descriptions}\label{sec:appendix_tasks}
The benchmark is organized into two complementary splits. The \textbf{main split} comprises VQA sets curated under diverse scenario settings and node-relationship configurations, spanning normal tabular data, time-series data, and spatial data. The \textbf{robustness split} reuses the same question--answer pairs from the main split but applies controlled visual perturbations to the images, ensuring that answer semantics and human readability remain intact so that any accuracy change can be attributed solely to the perturbation.

\paragraph{Main subset.}

The main subset is categorized by the underlying scenario and the linkage functions that connect nodes in the latent dependency graph. We curate five categories:
\begin{itemize}
    \item \textbf{SingleDense:} VQA sets built from a single chart type that contains a high volume of data points, stressing dense-context reading and fine-grained value extraction.
    \item \textbf{MultiChart:} The base category in which each VQA set combines heterogeneous chart types, requiring cross-chart aggregation and comparison.
    \item \textbf{StatAnalysis:} Sets focused on statistical measurement and distributional characteristics, featuring plots such as box plots, range plots, and histograms, with questions targeting mean, standard deviation, and other summary statistics.
    \item \textbf{TimeSequence:} Sets that introduce a specialized \textsc{Shift} function to encode temporal differences along the time axis, evaluating models' ability to track evolution and detect temporal trends.
    \item \textbf{Coordinates:} Sets that capture geometric and spatial relationships among multiple subjects, testing coordinate-aware and positional reasoning.
\end{itemize}
These categories reflect distinct real-world scenarios and structural settings. To ensure fair comparison, difficulty level, data volume, and question type are distributed independently and evenly across categories.

\paragraph{Robustness subset.}

To measure the stability of MLLMs under realistic visual degradation, we apply four families of perturbation to images from the main track without altering the underlying answers or human readability:
\begin{itemize}
    \item \textbf{Noise:} Image-wise noise, such as Gaussian noise, text-level noise, such as watermarks, and data-level noise, such as misleading background text printed alongside the chart.
    \item \textbf{Resolution:} Degradation through blurring, rescaling, and lossy compression, PNG-to-JPEG transformation.
    \item \textbf{Style Transfer:} Redrawing charts with a generative image model to emulate non-computer-generated styles, enabling assessment of whether performance shifts when charts depart from standard rendering templates.
\end{itemize}
By holding the question--answer pairs fixed, any performance variation in the robustness split can be directly attributed to the applied perturbation.

\section{Data Curation Theory}\label{sec:appendix_curation_theory}

The latent graph-based data dependencies is critical to ensure the consistency and uniqueness of a multi-chart set. This section outlines the construction process of long-chart VQA set, in which each question–answer pair is explicitly traceable to data nodes and visual evidence.

\subsection{Latent Dependency Graph}

The latent structure is represented by a directed acyclic graph (DAG),
\[
G = (V,E)
\]
where $V$ is the set of nodes and $E \subseteq V \times V$ is the set of edges. The node set is partitioned into three types of nodes,
\[
V = D \cup Y \cup I.
\]

\paragraph{Data nodes.} The root layer consists of source data nodes,
\[
D = \{D_1, \dots, D_{n_d}\},
\]
where each $D_j$ denotes an independent and non-redundant raw data object. These nodes have no parents in the graph, so that
\[
\operatorname{Pa}(D_j) = \varnothing, \qquad j=1,\dots,n_d.
\]

\paragraph{Computation nodes.} The computation node consists of derived variables,
\[
Y = \{Y_1, \dots, Y_{n_y}\},
\]
where each node $Y_k$ is obtained as a deterministic mathematical transformation of its parent set:
\[
Y_k = f_k\bigl(\operatorname{Pa}(Y_k)\bigr), \qquad k=1,\dots,n_y.
\]
Here, $f_k$ denotes a deterministic function and $\operatorname{Pa}(Y_k)$ is the set of immediate predecessors of $Y_k$ in $G$.

\paragraph{Image nodes.} The image node consists of visual representations,
\[
I = \{I_1, \dots, I_{n_i}\},
\]
where each visualization node is generated by a mapping from data or derived variables into a visual space:
\[
I_\ell = \phi_\ell\bigl(\operatorname{Pa}(I_\ell); \theta_\ell\bigr), \qquad \ell=1,\dots,n_i.
\]
In this expression, $\phi_\ell$ is a visualization map and $\theta_\ell$ denotes the specifies aesthetic parameters such as scale, layout, color, or geometry.

\subsection{Edges Mapping Information Flow}

Each directed edge in the network has an injectivity indicator, which characterizes the information-preserving capacity of the corresponding transformation. Let $u,v \in V$ and suppose that an edge $u \to v$ corresponds to an information flow from $u$ into $v$. We distinguish the following cases.

\paragraph{Bidirected Edges.}

An edge is \emph{bidirected}, denoted informally by
\[
u \leftrightarrow v,
\]
if the transformation is one-to-one and reversible. In this case, $u$ and $v$ are informationally equivalent, and the conditional entropies satisfy
\[
H(u \mid v) = 0 \qquad \text{and} \qquad H(v \mid u) = 0.
\]

\paragraph{Directed Edges.}

An edge is \emph{directed}, denoted by
\[
u \xrightarrow{\mathrm{inj}} v,
\]
if the target node $v$ can be transformed from $u$, although $v$ in reverse, cannot reconstruct $u$. Equivalently,
\[
H(u \mid v) = 0.
\]

\subsection{The Minimal Set and Subgraph Theory}

The latent graph design is introduced to satisfy the following objectives. First, all data are inter-related and consistent, ensuring that questions from any data node yield a unique answer. Second, the final set of images must collectively encode all information required to answer the complete set of questions. To formalize these objectives, we introduce the following concept for dataset curation.

\paragraph{Information equivalence classes.}

Define a relation $\sim$ on $V$ by declaring
\[
u \sim v
\]
if and only if there exists a path from $u$ to $v$ consisting only of bidirected transformations. Equivalently, $u$ and $v$ are \emph{informationally equivalent}, written $u \equiv v$, whenever each is recoverable from the other with zero information loss. In entropic terms, this implies
\[
H(u \mid v) = 0 \qquad \text{and} \qquad H(v \mid u) = 0.
\]

The equivalence classes induced by $\sim$ partition the vertex set $V$ into maximal subsets of mutually information-equivalent nodes. We denote the class containing $u$ by
\[
[u] = \{v \in V : v \sim u\}.
\]
Within such a class, any node serves as a perfect proxy for any other, regardless of whether it represents raw data, a derived variable, or a visualization. Two nodes connected by bidirected edges form a trivial equivalence class. For example, an image node bijectively linked to a data node contains all information necessary to answer any question associated with the class.

\subsection{The Information Basis}

We seek a subset
\[
S \subseteq V
\]
that acts as a minimal generative set for the full framework. Intuitively, $S$ is an information basis if it contains exactly the information required to reconstruct every node in the network, while excluding redundant or conflicting specifications.

Formally, $S$ is required to satisfy the following properties.

\paragraph{Completeness.}
The set $S$ is complete if it spans the entire framework in the sense that every node in $V$ is recoverable from $S$:
\[
H(V \mid S) = 0.
\]
Equivalently, the joint state of all nodes in the graph is fully determined once the elements of $S$ are known.

\paragraph{Independence.}
The set $S$ is independent if no element of $S$ is informationally redundant relative to the others. That is, for every $s_i \in S$,
\[
H\bigl(s_i \mid S \setminus \{s_i\}\bigr) > 0.
\]
Thus, removal of any element from $S$ causes a genuine loss of information.

\paragraph{Conflict-free structure.}
The set $S$ is conflict-free if it does not contain multiple nodes that encode the same information that potentially cause conflicts in data. Two important forbidden patterns are:
\begin{enumerate}
    \item selecting more than one node from the same information equivalence class; and
    \item selecting both a parent node and a deterministic descendant whose value is already fixed by that parent through the structural maps $f_k$ or $\phi_\ell$.
\end{enumerate}
The first introduces redundancy, while the second yields an overdetermined system vulnerable to inconsistency if supplied values deviate from deterministic transformations.

Finally, a \emph{minimal generative set} is a subset $S \subseteq V$ satisfying completeness, independence, and conflict-freeness simultaneously.

\subsection{Long-Chart VQA Set Curation Process}

Building on the above theory, we design a structured curation pipeline. The synthetic process begins by manually defining the latent dependency graph and uses algorithm to identify a minimal generative set within it. Data in this minimal set is then synthesized by LLM according to human-specified requirements. Subsequently, the rest of data or computation node on the the graph is propagated, which ensure all data and images are global consistent.

After data curation is complete, questions are generated from data nodes, while images are produced from visualization nodes in parallel. Finally, the process performs a backward propagation step to verify sufficiency to answer each question. An algorithm verify that curated images along collectively form a minimal information set, thereby guaranteeing that all questions can be answered solely from the provided charts. Finally, a human verification will examine all questions are answerable, answer are accurate and images are aesthetically well-built.

\subsection{Global Data Structures}

The implementation is organized around a small collection of persistent objects.

\paragraph{Graph specification.}
Each VQA-set contains human-specified graph structure initially. The graph is an abstracted dependency graph some real-life scenarios. The graph specification stores the node set, edge set, node types, structural equations, and attribute schemas:
\[
\mathsf{GraphSpec} = (V,E,\mathcal{T},\mathcal{E}).
\]
Here, $\mathcal{T}$ records node types, $\mathcal{E}$ records the operator and information flow attributes between data nodes.

\paragraph{Minimal-set result.}
The algorithm of graph traversal provide the following results
\[
\mathsf{MinimalSetResult} = (S^{\ast},\mathcal{S}_{\mathrm{alt}}),
\]
where $S^{\ast}$ is the unique selected minimal set, $\mathcal{S}_{\mathrm{alt}}$ is the set of admissible alternatives. 

\paragraph{Requirement Set.}
Requirements in data synthesis are necessary to ensure that the resulting base data is meaningful. When drafting the latent graph, requirements are specified by humans to guide the simulation of data in the minimal set. Formally, requirements are stored as

\[
\mathsf{RequirementSet} = \{r_1,\dots,r_{n}\},
\]

where each $r_j$ denotes a seperate text instruction that input to LLM at the data synthesis stage. 

\paragraph{Data Synthesis.}
Node data are generated based on the graph specification and the requirement set. 

For data nodes without parents, values are synthesized directly by the LLM according to the requirements:

\[
D_d = f_{LLM}(\mathsf{GraphSpec}, \mathsf{RequirementSet}).
\]

For computation nodes with parents, values are propagated using the predefined operators encoded in the edge connections:

\[
D_c = f_{propagate}(\mathsf{GraphSpec}, D_d).
\]

Finally, all node data are collected into a unified set, which is then passed in parallel to chart generation and QA generation:

\[
\mathsf{NodeData} = \{D_d, D_c\}.
\]

\paragraph{VQA Generation.}
Once data are synthesized, question–answer pairs are generated from the node data. Questions are instructed to be evenly distributed across data nodes and computation nodes, ensuring a balanced difficulty profile across computational complexity:

\[
\mathsf{QA} = f_{LLM}(\mathsf{GraphSpec}, \mathsf{NodeData}).
\]

Candidate QA sets are then manually validated. In parallel, images are generated from the same node data:

\[
\mathsf{Images} = f_{LLM}(\mathsf{GraphSpec}, \mathsf{NodeData}).
\]

\paragraph{Final dataset package.}
The output of the full pipeline is
\[
\mathsf{Dataset} = (\mathsf{GraphSpec},\mathsf{NodeData},\mathsf{Images}, \newline
\mathsf{QA}),
\]
containing the graph, generated data, images, selected question-answer pairs, and all meta information indicating the question characteristics.

\subsection{Stepwise Dataset Curation Pipeline}

Building on the curation techniques described above, we present the long-chart VQA curation pipeline, summarized in Algorithm~\ref{code: curation}.

\begin{algorithm*}[t]
\caption{Pseudocode for the VQA Set Curation Pipeline}
\label{code: curation}
\begin{algorithmic}[1]
\Require Specifed latent graph $\mathsf{GraphSpec}$, simulation requirements $\mathcal{R}$.
\Ensure Long-Chart VQA Curation 
\State $\mathsf{MinimalSet} \leftarrow f_{traversal}(\mathsf{GraphSpec})$
\State $\mathsf{DataMinimalSet} \leftarrow f_{LLM}(\mathsf{GraphSpec}, R)$
\State $\mathsf{NodeData} \leftarrow f_{propagate}(\mathsf{GraphSpec}, \mathsf{DataMinimalSet})$
\State $\mathsf{QACandidates}\leftarrow f_{LLM}(\mathsf{GraphSpec}, \mathsf{NodeData})$
\State $\mathsf{Images} \leftarrow \textsc(\text{RenderImages})(\mathsf{GraphSpec}, \mathsf{NodeData})$
\State $\mathsf{DependencyGraph} \leftarrow \textsc{UpdateGraph}(\mathsf{GraphSpec}, \mathsf{QACandidates}, \mathsf{Images})$
\Comment{{\color{gray}Verify images can sufficiently support each question; assign structural difficulty $\operatorname{hop}(q)$}}
\State $\mathsf{VerifiedVQA} \leftarrow \textsc{HumanVerify}(\mathsf{GraphSpec}, \mathsf{NodeData}, \mathsf{Images}, \mathsf{QACandidates})$
\Comment{{\color{gray}Human-in-the-loop: confirm readability, answer uniqueness, coherent easy-to-hard path, and analytical validity}}
\end{algorithmic}
\end{algorithm*}

\section{Annotation and Quality Control}
\label{sec:appendix_annotation}

We developed a dedicated annotation interface to support efficient QA construction and systematic quality control. The interface allows annotators to inspect candidate images, write and revise question--answer pairs, and verify whether each item is fully grounded in the provided visual evidence. This setup helps standardize the annotation workflow, reduce formatting inconsistencies, and make it easier to trace potential errors during the verification stage.

\begin{figure}[H]
\centering
\includegraphics[width=0.92\linewidth]{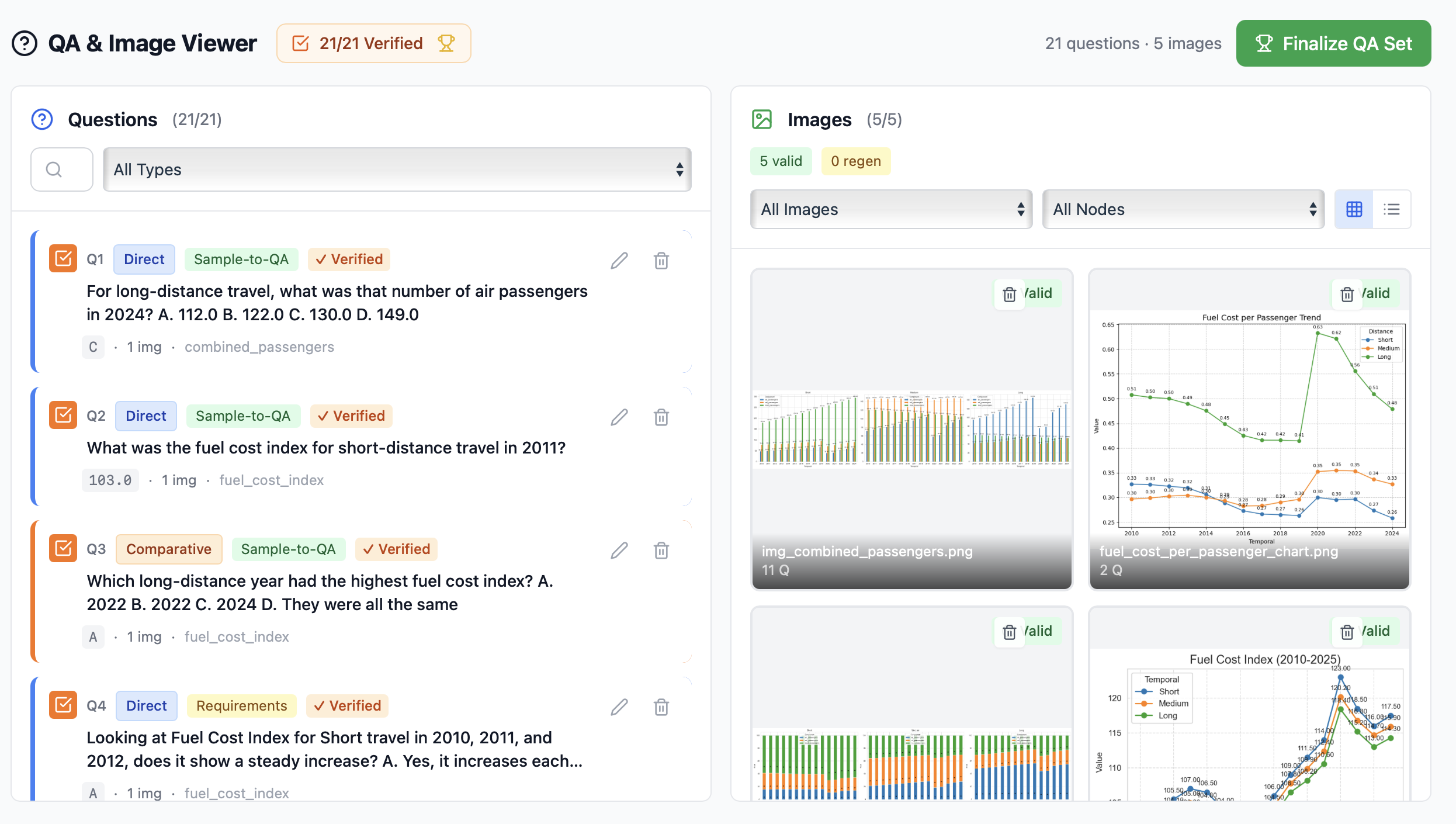}
\caption{Annotation interface developed to annotate and verify the QA \& images.}
\label{fig:appendix_labeling}
\end{figure}

\section{Examples of VQA Sets}
\label{sec:appendix_examples}
\Needspace{6\baselineskip}
\subsection{VQA Set Sample 1}\label{subsec:sample_1}
This sample VQA set includes diverse image types such as pie charts, line charts, bar charts, and tables. It represents a relatively complex case, containing more than 200 data points.

\Needspace{6\baselineskip}
\subsection{VQA Set Sample 2}\label{subsec:sample_2}
This sample VQA set features stacked bar charts, histograms, range plots, and box plots, illustrating varied chart structures.

\Needspace{6\baselineskip}
\subsection{VQA Set Sample 3}\label{subsec:sample_3}
This sample demonstrates image style transfer. Line plots and tables are rendered in a whiteboard style while preserving the readability of all data and information.

\Needspace{6\baselineskip}
\subsection{VQA Set Sample 4}\label{subsec:sample_4}
This sample illustrates text noise injection. Watermarks are applied to all images while maintaining readability, ensuring that the underlying data and information remain accessible.

\begin{figure*}[tbp]
\centering
\begin{minipage}[t]{0.47\textwidth}
  \centering
  \includegraphics[width=0.78\linewidth]{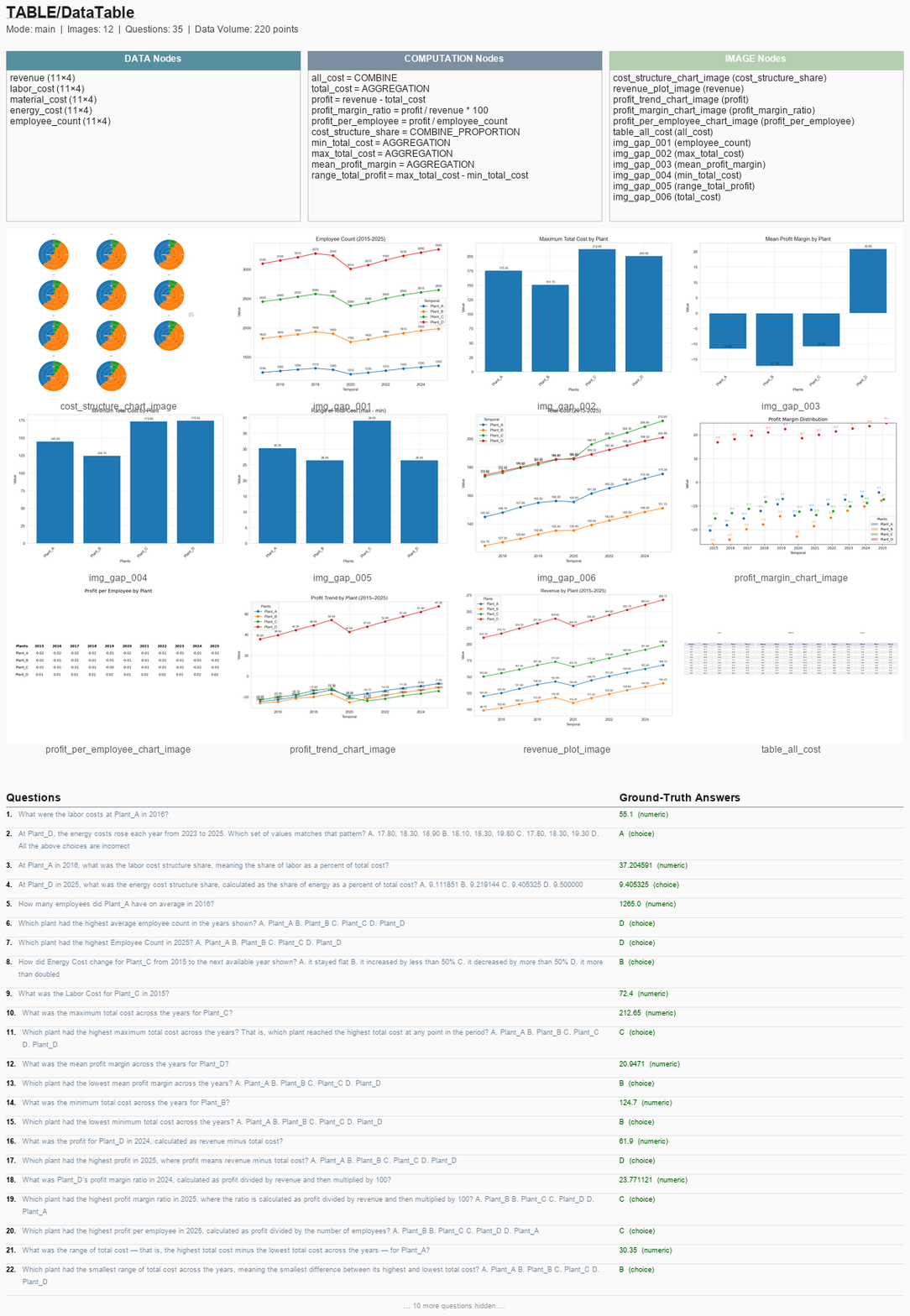}
  \small (a) Sample VQA set 1.
\end{minipage}\hfill
\begin{minipage}[t]{0.47\textwidth}
  \centering
  \includegraphics[width=0.78\linewidth]{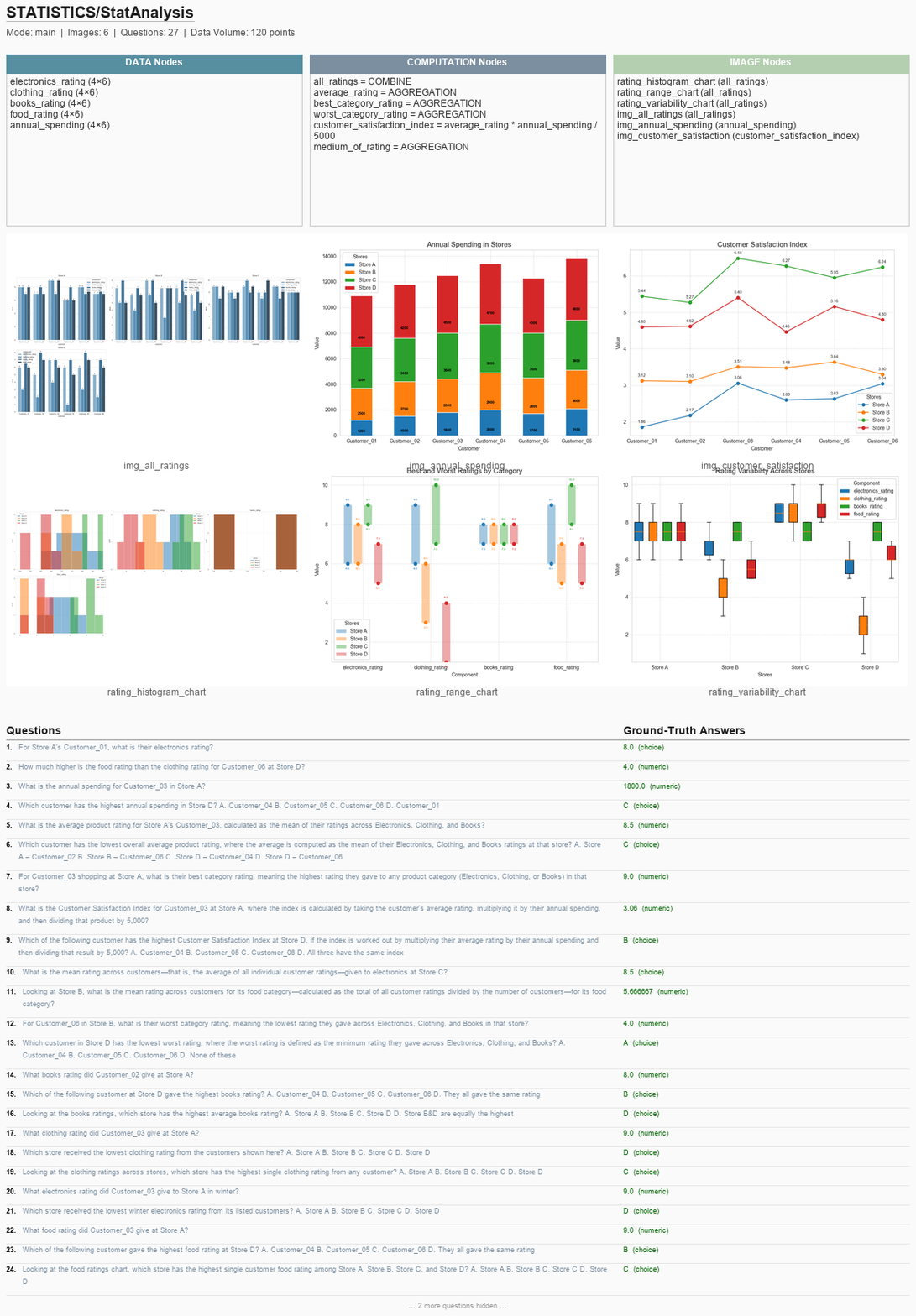}
  \small (b) Sample VQA set 2.
\end{minipage}

\vspace{4pt}

\begin{minipage}[t]{0.47\textwidth}
  \centering
  \includegraphics[width=0.78\linewidth]{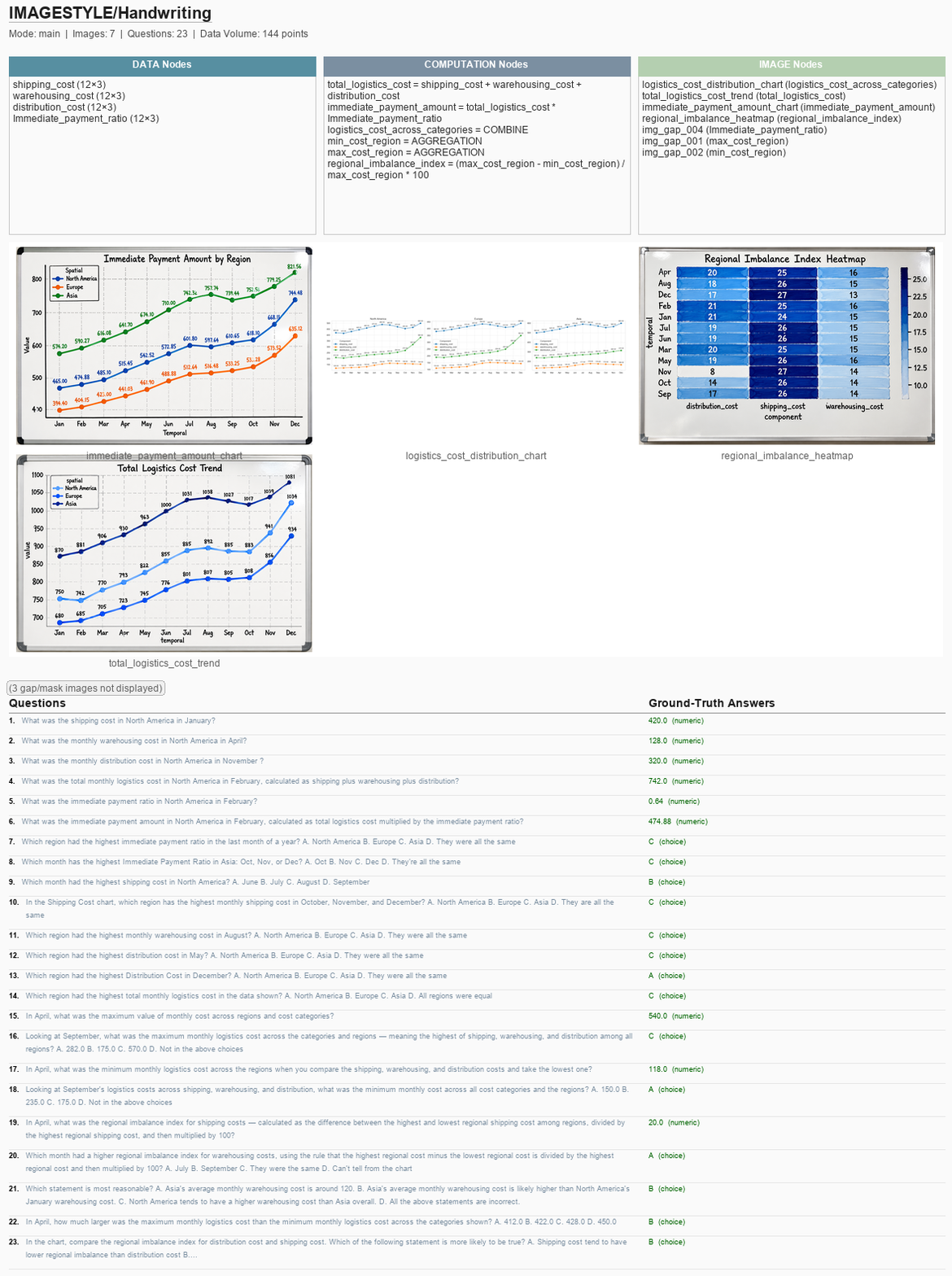}
  \small (c) Sample VQA set 3.
\end{minipage}\hfill
\begin{minipage}[t]{0.47\textwidth}
  \centering
  \includegraphics[width=0.78\linewidth]{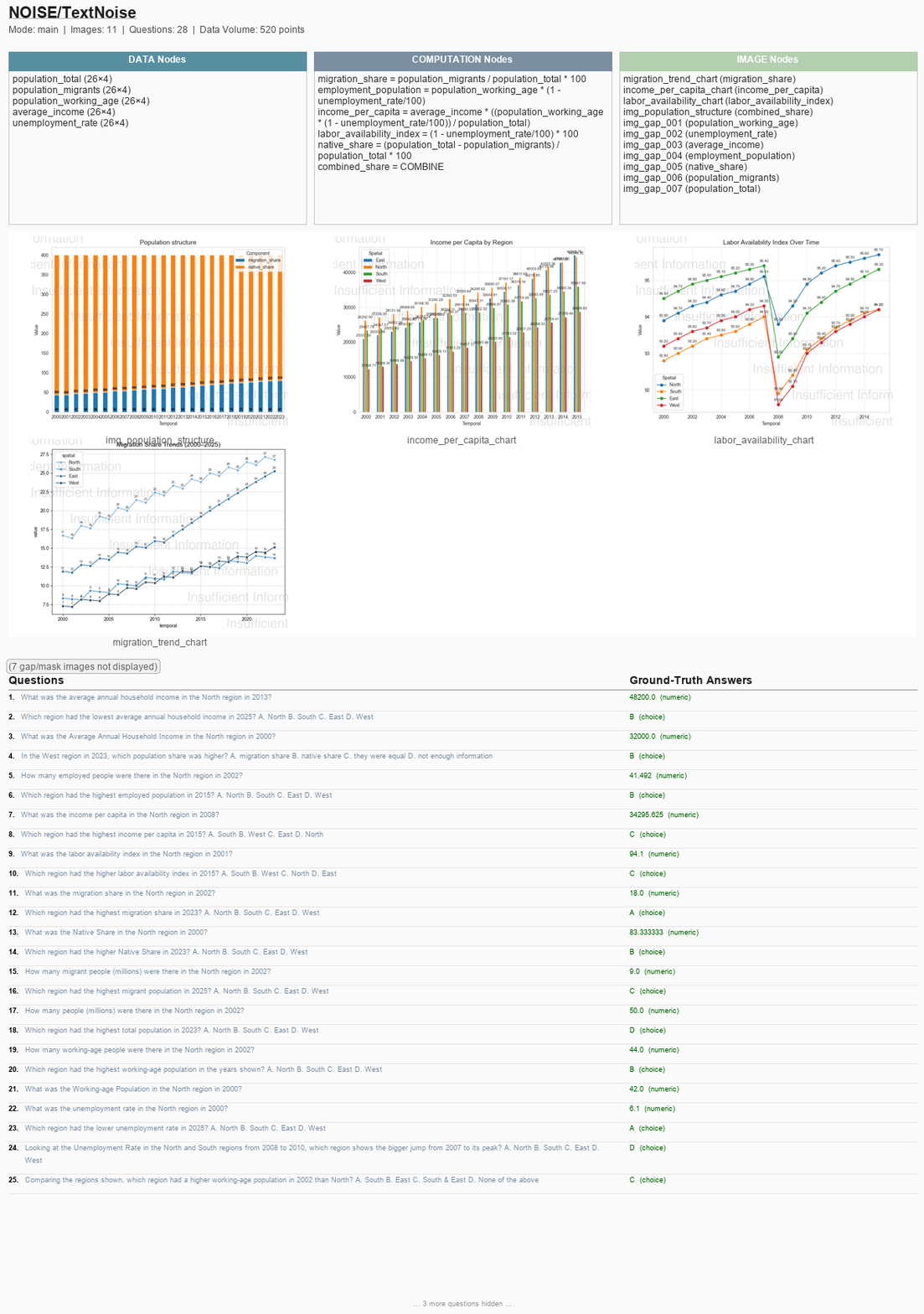}
  \small (d) Sample VQA set 4.
\end{minipage}
\caption{Examples of VQA sets covering mixed chart types, statistical views, style transfer, and text-noise perturbation. These figures may need to enlarge to read.}
\label{fig:appendix_vqa_samples}
\end{figure*}

\section{Evaluation Metrics}
\label{sec:evaluation_metrics}

\subsection{Accuracy}
The accuracy metric measures the proportion of questions answered correctly:
\[
\text{Accuracy} = \frac{\sum_{i=1}^{N} \mathbb{I}\bigl(\hat{a}_i = a_i\bigr)}{N},
\]
where $N$ is the total number of questions, $\hat{a}_i$ is the predicted answer for the $i$-th question, $a_i$ is the ground-truth answer, and $\mathbb{I}(\cdot)$ is the indicator function that returns 1 when the prediction matches the ground truth and 0 otherwise.

The exact matching criterion depends on the question type:
\begin{itemize}
    \item \textbf{Multiple-choice questions.} An answer is regarded as correct only if the selected option exactly matches the ground-truth option (e.g., \texttt{A}, \texttt{B}, \texttt{C}, or \texttt{D}).
    \item \textbf{Numerical questions.} An answer is regarded as correct if the relative error is within 1\% of the standard answer, i.e., $\frac{|\hat{a}_i - a_i|}{|a_i|} \le 0.01$.
\end{itemize}

\subsection{mDIoU: Modified Distance IoU for Grounding Boxes}
For evaluating bounding-box grounding quality, we adopt a distance-aware intersection-over-union variant. Let $\hat{B}$ denote the predicted bounding box and $B$ the ground-truth box. The modified Distance IoU (mDIoU) is defined as
\[
\text{mDIoU}(\hat{B}, B) = \text{IoU}(\hat{B}, B) - \frac{\rho^2\bigl(c_{\hat{B}}, c_B\bigr)}{d^2},
\]
where $\text{IoU}(\hat{B}, B)$ is the standard intersection-over-union of the two boxes, $c_{\hat{B}}$ and $c_B$ are the center points of the predicted and ground-truth boxes, $\rho(\cdot, \cdot)$ is the Euclidean distance between centers, and $d$ is the diagonal length of the smallest enclosing box that contains both $\hat{B}$ and $B$. The second term penalizes predictions that are correctly sized but spatially displaced.

We choose DIoU over plain IoU because model-generated bounding boxes are often much larger than the ground-truth regions; in such cases the intersection area is small relative to the union, yielding a near-zero IoU that is hard to discriminate across models. By adding a center-distance penalty, DIoU rewards predictions whose centers are close to the ground-truth center even when the box sizes differ, making the metric more comparable across systems. The metric is implemented from scratch.

\subsection{Chamfer Distance for Point Sets}
For evaluating point-level grounding, we measure the bidirectional Chamfer distance between the predicted point set $\hat{\mathcal{P}} = \{\hat{p}_1, \dots, \hat{p}_m\}$ and the ground-truth point set $\mathcal{P} = \{p_1, \dots, p_n\}$:
\[
\text{CD}(\hat{\mathcal{P}}, \mathcal{P}) = \frac{1}{m}\sum_{\hat{p}\in\hat{\mathcal{P}}} \min_{p\in\mathcal{P}} \|\hat{p} - p\|_2 \;+\; \frac{1}{n}\sum_{p\in\mathcal{P}} \min_{\hat{p}\in\hat{\mathcal{P}}} \|\hat{p} - p\|_2.
\]
The first term averages the nearest-neighbor distance from each predicted point to the ground-truth set, and the second term averages the nearest-neighbor distance from each ground-truth point to the predicted set. This symmetric formulation penalizes both false-positive predictions and missed ground-truth points. The metric is implemented from scratch. 


\section{Baseline Configurations}
\label{sec:baseline_config}
\begin{table}[H]
\centering
\caption{\textbf{main context}}
\label{tab:baseline_config}
\small
\setlength{\tabcolsep}{3pt}
\resizebox{\linewidth}{!}{%
\begin{tabular}{lcccc}
\toprule
\textbf{Model} & \textbf{Release Date} & \textbf{Max Tokens} & \textbf{Temperature} & \textbf{Tok-P} \\
\midrule
doubao-seed-2.0-lite & 2026-02-25 & 10000 & 0 & 1.0 \\
doubao-seed-2.0-pro & 2026-02-25 & 10000 & 0 & 1.0 \\
gemini-3.1-flash-lite & 2026-02-20 & 10000 & 0 & 1.0 \\
gpt-5.4-mini & 2026-03-18 & 10000 & 0 & 1.0 \\
gpt-5.4-nano & 2026-03-18 & 10000 & 0 & 1.0 \\
qwen3.5-flash & 2026-03-04 & 10000 & 0 & 1.0 \\
\midrule
gemma-4-26b-a4b-it & 2026-04-04 & 10000 & 0 & 1.0 \\
gemma-4-31b-it & 2026-04-04 & 10000 & 0 & 1.0 \\
mimo-v2.5 & 2026-04-23 & 10000 & 0 & 1.0 \\
qwen3.5-35b-a3b & 2026-03-04 & 10000 & 0 & 1.0 \\
claude-sonnet-4-5 & 2025-09-30 & 10000 & 0 & 1.0 \\
\bottomrule
\end{tabular}%
}
\end{table}

\onecolumn

\section{Prompts}
\label{sec:prompts}

\subsection{Baseline Assessment Prompts}
\Needspace{12\baselineskip}
\textbf{Prompts for baseline assessment}
\begin{promptframe}
Please carefully examine the images provided above. They contain charts, tables, or data visualizations. Based on the information shown in these images, answer the following questions one by one. For each question, respond with a JSON object containing 'LLM\_answer'. The final answer: a number, A/B/C/D, or True/False and 'reasoning' is a brief explanation of your reasoning. 
\end{promptframe}

Input charts and provides the following per-question prompt, appended sequentially for each question: 

\begin{promptframe}
Question \{idx\}: \{question\_text\}  

Options:\{options\_text\}

Please answer in the following JSON format exactly: 

\{ 

"LLM\_answer\_initial": "your initial answer here (a number, A/B/C/D, or True/False)",  

"reasoning": "brief explanation of how you arrived at this answer",

"LLM\_answer\_final": "your final answer here (a number, A/B/C/D, or True/False)" 

\}                                                                                 
\end{promptframe}

\Needspace{12\baselineskip}
\textbf{Prompts for the retrieval mode}
\begin{promptframe}
 Please carefully examine the images provided above. They contain charts, tables, or data visualizations.Your first task is to reconstruct ALL the underlying data shown in these images as completely and accurately as possible. For each chart or visualization, extract:  
\begin{promptitemize}
    \item The chart title 
    \item All axis labels, categories, and legends 
    \item All numerical values with their corresponding row/column/category labels 
\end{promptitemize}

Present the reconstructed data in a clear structured format (JSON or markdown table). This reconstruction will be used to answer subsequent questions. 
\end{promptframe}

\Needspace{12\baselineskip}
\textbf{Prompts for the grounding mode}
\begin{promptframe}
 You are answering questions about chart images. For each question, provide your best answer AND identify the rectangular regions in each image that support your answer. 
  
\textbf{Coordinate convention:}

\begin{promptitemize}
    \item Use pixel coordinates with top-left origin: (0, 0) is the top-left pixel.
    \item x increases rightward; y increases downward. 
    \item Each bbox is [x\_min, y\_min, x\_max, y\_max] as integers. 
    \item Legal ranges: x in [0, {width\_px}]; y in [0, \{height\_px\}]. 
\end{promptitemize}

  \textbf{Image labels:}
  
  Each image below is labeled with its ID and title. Use the exact image\_id in your output. 
  
\{image\_descriptions\}  
  
\textbf{Instructions:} 

\begin{promptitemize}
    \item First, give your initial answer and the bounding boxes that support it.
    \item Then explain your reasoning.
    \item Finally, give your final answer and revised bounding boxes (they may be the same as initial).
\end{promptitemize}

If an image is NOT relevant to the answer, you MUST include it with an empty bbox list: "bboxes": []. Do not omit any image. 

Respond in this exact JSON format:
  
  \{
  
  "initial\_answer": "your initial answer (number, A/B/C/D, or True/False)",
    
    "initial\_bbox": \{
    
    "{example\_image\_id}": \{"bboxes": [[x\_min, y\_min, x\_max, y\_max], ...]\}
    
      \},  
    
    "reasoning": "brief explanation of how you arrived at this answer and why you selected these regions", 
    
    "final\_answer": "your final answer (number, A/B/C/D, or True/False)", 
    
    "final\_bbox": \{"{example\_image\_id}": {"bboxes": [[x\_min, y\_min, x\_max, y\_max], ...]}
      \} 
   
\}  
\end{promptframe}

\Needspace{12\baselineskip}
\textbf{Prompts for the point mode}
\begin{promptframe}
 You are answering questions about chart images. For each question, provide your best answer AND identify the exact point locations in each image that support your answer.    
 
 \textbf{Coordinate convention:}
 \begin{promptitemize}
     \item Use pixel coordinates with top-left origin: (0, 0) is the top-left pixel.
     \item x increases rightward; y increases downward. 
     \item Each point is [x, y] as integers.  
     \item Legal ranges: x in [0, {width\_px}]; y in [0, {height\_px}].  
 \end{promptitemize}

\textbf{Image labels:}  
Each image below is labeled with its ID and title. Use the exact image\_id in your output. 

\{image\_descriptions\}  
  
\textbf{Instructions:} 
\begin{promptitemize}
    \item First, give your initial answer and the point coordinates that support it.
    \item Then explain your reasoning.    
    \item Finally, give your final answer and revised point coordinates (they may be the same as initial). 
\end{promptitemize}

For each relevant data value, provide ONE point coordinate that best represents its location in the image. For bar charts, this is typically the center or top of the bar. For line charts, this is the data point marker. For pie charts, this is the center of the slice.

If an image is NOT relevant to the answer, you MUST include it with an empty point list: "points": []. Do not omit any image. 

Respond in this exact JSON format: 

\{                                       
    "initial\_answer": "your initial answer (number, A/B/C/D, or True/False)",
    
    "initial\_points": \{  
    "{example\_image\_id}": {"points": [[x, y], ...]}  
    \},  
    
    "reasoning": "brief explanation of how you arrived at this answer and why you selected these points",                 
    
    "final\_answer": "your final answer (number, A/B/C/D, or True/False)",
    
    "final\_points": \{ 
    "{example\_image\_id}": {"points": [[x, y], ...]}  
    \}  
    
\}  
\end{promptframe}

%% file: tables/appendix_taxonomy_view1.tex
\begin{table*}[tbp]
\centering
\caption{\textbf{MLLM performance across VQA curation categories under Extended Context.} The best-performing model in each category is typeset in \textbf{boldface}, and the runner-up is indicated with \textit{underline}.}
\label{tab:appendix_taxonomy_view1}
\setlength{\tabcolsep}{4pt}
\resizebox{\textwidth}{!}{%
\begin{tabular}{lcccccccccc}
\toprule
\multirow{2}{*}{\textbf{Model}} & \multicolumn{6}{c}{\textbf{Main}} & \multicolumn{4}{c}{\textbf{Robustness}} \\
\cmidrule(lr){2-7} \cmidrule(lr){8-11}
& \textbf{GEOMETRY} & \textbf{MULTICHART} & \textbf{STATISTICS} & \textbf{TABLE} & \textbf{TIMESEQ} & \textbf{Total} & \textbf{IMAGESTYLE} & \textbf{NOISE} & \textbf{RESOLUTION} & \textbf{Total} \\
\midrule
doubao-seed-2.0-pro & \textbf{90.34\%} & \textbf{84.64\%} & \textbf{79.25\%} & \textbf{93.13\%} & \textbf{66.86\%} & \textbf{84.15\%} & \textbf{85.37\%} & \textbf{86.99\%} & \textbf{87.46\%} & \textbf{86.88\%} \\
doubao-seed-2.0-lite & \textit{\underline{86.90\%}} & \textit{\underline{83.70\%}} & \textit{\underline{77.36\%}} & \textit{\underline{90.15\%}} & \textbf{66.86\%} & \textit{\underline{82.29\%}} & \textit{\underline{80.49\%}} & \textit{\underline{84.09\%}} & \textit{\underline{85.71\%}} & \textit{\underline{84.03\%}} \\
gemini-3.1-flash-lite-preview & 79.31\% & \textbf{84.64\%} & 71.70\% & 86.57\% & 61.14\% & 78.75\% & 80.49\% & 80.60\% & 78.22\% & 79.79\% \\
gemma-4-31b-it & 20.00\% & 56.74\% & 41.98\% & 28.66\% & 1.71\% & 33.56\% & 42.59\% & 40.28\% & 49.20\% & 43.85\% \\
gpt-5.4-mini & 73.79\% & 73.35\% & 61.32\% & 82.39\% & 53.71\% & 70.91\% & 70.73\% & 74.22\% & 74.91\% & 73.87\% \\
gpt-5.4-nano & 48.28\% & 57.05\% & 52.83\% & 61.79\% & 28.00\% & 52.28\% & 53.31\% & 46.11\% & 53.48\% & 49.77\% \\
qwen3.5-35b-a3b & 86.21\% & 74.92\% & 75.00\% & 88.06\% & 56.57\% & 77.32\% & 78.05\% & 79.21\% & 81.18\% & 79.67\% \\
qwen3.5-flash & 83.45\% & 77.43\% & 74.06\% & 82.09\% & 55.43\% & 75.63\% & 74.91\% & 77.35\% & 75.78\% & 76.42\% \\
\bottomrule
\end{tabular}%
}
\end{table*}

%% file: tables/appendix_taxonomy_view2.tex
\begin{table*}[tbp]
\centering
\caption{\textbf{MLLM performance across VQA curation categories under In-Depth evaluation.} The best-performing model in each category is typeset in \textbf{boldface}, and the runner-up is indicated with \textit{underline}.}
\label{tab:appendix_taxonomy_view2}
\setlength{\tabcolsep}{4pt}
\resizebox{\textwidth}{!}{%
\begin{tabular}{lcccccccccc}
\toprule
\multirow{2}{*}{\textbf{Model}} & \multicolumn{6}{c}{\textbf{Main}} & \multicolumn{4}{c}{\textbf{Robustness}} \\
\cmidrule(lr){2-7} \cmidrule(lr){8-11}
 & \textbf{GEOMETRY} & \textbf{MULTICHART} & \textbf{STATISTICS} & \textbf{TABLE} & \textbf{TIMESEQ} & \textbf{Total} & \textbf{IMAGESTYLE} & \textbf{NOISE} & \textbf{RESOLUTION} & \textbf{Total} \\
\midrule
doubao-seed-2.0-lite & \textit{\underline{67.59\%}} & \textit{\underline{66.46\%}} & \textit{\underline{64.62\%}} & 77.31\% & \textit{\underline{42.29\%}} & \textit{\underline{65.77\%}} & 61.54\% & 77.49\% & \textit{\underline{82.39\%}} & 77.15\% \\
doubao-seed-2.0-pro & \textbf{68.97\%} & \textbf{72.41\%} & \textbf{66.51\%} & \textbf{82.99\%} & \textbf{49.71\%} & \textbf{70.57\%} & \textbf{86.76\%} & \textbf{88.73\%} & \textbf{89.20\%} & \textbf{88.56\%} \\
gemma-4-31b-it & 57.24\% & 61.13\% & 42.45\% & 46.27\% & 22.86\% & 47.47\% & 39.72\% & 37.05\% & 43.90\% & 39.78\% \\
gpt-5.4-mini & 56.55\% & 61.76\% & 62.74\% & 71.64\% & 33.71\% & 59.95\% & 70.73\% & 72.36\% & 73.34\% & 72.42\% \\
gpt-5.4-nano & 46.21\% & 49.53\% & 34.43\% & 54.03\% & 20.57\% & 43.42\% & 50.17\% & 47.27\% & 52.96\% & 49.65\% \\
mimo-v2.5 & 60.00\% & 66.14\% & 59.43\% & 76.12\% & 32.00\% & 61.97\% & 77.35\% & 76.77\% & 75.44\% & 76.42\% \\
qwen3.5-35b-a3b & 61.38\% & 57.05\% & 60.85\% & 77.91\% & 41.14\% & 61.80\% & \textit{\underline{78.75\%}} & \textit{\underline{79.33\%}} & 78.40\% & \textit{\underline{78.92\%}} \\
qwen3.5-flash & 61.38\% & 58.31\% & 52.83\% & \textit{\underline{80.90\%}} & 36.00\% & 60.79\% & 67.94\% & 77.24\% & 79.09\% & 76.31\% \\
\bottomrule
\end{tabular}%
}
\end{table*}

%% file: tables/appendix_figure_type_breakdown.tex
\begin{table*}[tbp]
\centering
\caption{\textbf{MLLM performance across figure types under Extended Context.} The best-performing model in each category is typeset in \textbf{boldface}, and the runner-up is indicated with \textit{underline}.}
\label{tab:appendix_figure_type_breakdown_breadth}
\setlength{\tabcolsep}{3pt}
\resizebox{\textwidth}{!}{%
\begin{tabular}{lccccccccccc}
\toprule
\textbf{Model} & \textbf{Bar} & \textbf{Box plot} & \textbf{Heatmap} & \textbf{Histogram} & \textbf{Line} & \textbf{Map} & \textbf{Pie} & \textbf{Range plot} & \textbf{Scatter} & \textbf{Stacked bar} & \textbf{Table} \\
\midrule
doubao-seed-2.0-pro & \textbf{85.83\%} & \textbf{100.00\%} & 77.78\% & \textbf{96.43\%} & \textbf{83.77\%} & \textbf{74.29\%} & \textbf{85.00\%} & \textit{\underline{86.05\%}} & \textbf{66.67\%} & 72.73\% & \textbf{92.22\%} \\
doubao-seed-2.0-lite & \textit{\underline{85.43\%}} & 66.67\% & 77.78\% & 91.07\% & \textit{\underline{82.45\%}} & 71.43\% & \textbf{85.00\%} & \textbf{90.70\%} & 58.33\% & 54.55\% & \textit{\underline{85.56\%}} \\
gemini-3.1-flash-lite-preview & 81.10\% & \textbf{100.00\%} & 74.07\% & 82.14\% & \textit{\underline{82.45\%}} & 68.57\% & \textbf{85.00\%} & 76.74\% & 50.00\% & \textbf{90.91\%} & 83.33\% \\
gemma-4-31b-it & 40.14\% & 0.00\% & 46.67\% & 46.43\% & 37.06\% & 28.57\% & 50.00\% & 33.78\% & 20.83\% & 45.45\% & 14.29\% \\
gpt-5.4-mini & 71.26\% & 33.33\% & 62.96\% & 76.79\% & 72.19\% & 71.43\% & 75.00\% & 69.77\% & 58.33\% & 72.73\% & 77.78\% \\
gpt-5.4-nano & 51.97\% & 33.33\% & 48.15\% & 64.29\% & 53.64\% & 54.29\% & 50.00\% & 44.19\% & 12.50\% & 63.64\% & 62.22\% \\
mimo-v2.5 & 74.71\% & 33.33\% & 72.97\% & 91.07\% & 72.19\% & 66.67\% & 65.00\% & 76.60\% & 62.50\% & \textit{\underline{76.47\%}} & 77.48\% \\
qwen3.5-35b-a3b & 77.56\% & \textit{\underline{66.67\%}} & \textbf{81.48\%} & 91.07\% & 79.14\% & \textbf{74.29\%} & 80.00\% & 53.49\% & \textbf{66.67\%} & 63.64\% & 81.11\% \\
qwen3.5-flash & 75.59\% & 33.33\% & \textbf{81.48\%} & 91.07\% & 73.51\% & \textbf{74.29\%} & \textbf{85.00\%} & 72.09\% & \textbf{66.67\%} & 54.55\% & 80.00\% \\
\midrule
\textbf{Total} & 71.68\% & 51.85\% & 69.83\% & 79.64\% & 70.89\% & 65.76\% & 74.12\% & 67.53\% & 50.83\% & 65.35\% & 72.30\% \\
\bottomrule
\end{tabular}%
}
\end{table*}

\begin{table*}[tbp]
\centering
\caption{\textbf{MLLM performance across figure types under In-Depth evaluation.} The best-performing model in each category is typeset in \textbf{boldface}, and the runner-up is indicated with \textit{underline}.}
\label{tab:appendix_figure_type_breakdown_depth}
\setlength{\tabcolsep}{4pt}
\resizebox{\textwidth}{!}{%
\begin{tabular}{lccccccccc}
\toprule
\textbf{Model} & \textbf{Bar} & \textbf{Heatmap} & \textbf{Histogram} & \textbf{Line} & \textbf{Map} & \textbf{Pie} & \textbf{Range plot} & \textbf{Stacked bar} & \textbf{Table} \\
\midrule
doubao-seed-2.0-lite & 64.98\% & \textbf{65.91\%} & \textbf{56.45\%} & \textit{\underline{69.55\%}} & \textit{\underline{73.17\%}} & \textbf{71.43\%} & \textbf{59.26\%} & \textit{\underline{87.50\%}} & 67.84\% \\
doubao-seed-2.0-pro & \textbf{72.06\%} & \textit{\underline{63.64\%}} & \textit{\underline{53.23\%}} & \textbf{80.66\%} & \textbf{75.61\%} & \textit{\underline{66.67\%}} & \textit{\underline{29.41\%}} & \textbf{100.00\%} & \textbf{70.07\%} \\
gemma-4-31b-it & 44.12\% & 52.27\% & 24.19\% & 53.91\% & 70.73\% & 38.10\% & 17.65\% & 62.50\% & 42.86\% \\
gpt-5.4-mini & 56.37\% & 47.73\% & \textit{\underline{53.23\%}} & 64.20\% & 70.73\% & 61.90\% & \textit{\underline{29.41\%}} & 62.50\% & 57.82\% \\
gpt-5.4-nano & 44.12\% & 45.45\% & 17.74\% & 40.74\% & 60.98\% & 47.62\% & 17.65\% & 37.50\% & 49.66\% \\
mimo-v2.5 & 61.27\% & 52.27\% & 40.32\% & 66.67\% & 65.85\% & \textit{\underline{66.67\%}} & 23.53\% & 75.00\% & 66.67\% \\
qwen3.5-35b-a3b & \textit{\underline{69.12\%}} & 40.91\% & 46.77\% & 68.72\% & \textit{\underline{73.17\%}} & 38.10\% & 23.53\% & \textit{\underline{87.50\%}} & 63.27\% \\
qwen3.5-flash & 65.69\% & 59.09\% & 37.10\% & 65.84\% & 70.73\% & 61.90\% & \textit{\underline{29.41\%}} & \textbf{100.00\%} & \textit{\underline{68.03\%}} \\
\midrule
\textbf{Total} & 59.72\% & 53.41\% & 41.13\% & 63.79\% & 70.12\% & 56.55\% & 28.73\% & 76.56\% & 60.78\% \\
\bottomrule
\end{tabular}%
}
\end{table*}

%% file: tables/appendix_before_depth.tex
\begin{table*}[tbp]
\centering
\caption{\textbf{Performance before reasoning under Extended Context.} The best-performing model in each category is typeset in \textbf{boldface}, and the runner-up is indicated with \textit{underline}. An em dash (---) denotes unavailable results.}
\label{tab:appendix_before_breadth}
\setlength{\tabcolsep}{3pt}
\resizebox{\textwidth}{!}{%
\begin{tabular}{lcccccccccc}
\toprule
\multirow{2}{*}{\textbf{Model}} & \multicolumn{6}{c}{\textbf{Main}} & \multicolumn{4}{c}{\textbf{Robustness}} \\
\cmidrule(lr){2-7} \cmidrule(lr){8-11}
& \textbf{GEOMETRY} & \textbf{MULTICHART} & \textbf{STATISTICS} & \textbf{TABLE} & \textbf{TIMESEQ} & \textbf{Total} & \textbf{IMAGESTYLE} & \textbf{NOISE} & \textbf{RESOLUTION} & \textbf{Total} \\
\midrule
doubao-seed-2.0-pro & \textbf{90.34\%} & \textbf{84.64\%} & \textbf{78.30\%} & \textbf{92.84\%} & \textbf{66.86\%} & \textbf{85.37\%} & \textbf{86.99\%} & \textbf{87.46\%} & \textbf{83.90\%} & \textbf{86.88\%} \\
doubao-seed-2.0-lite & \textit{\underline{84.83\%}} & \textit{\underline{83.39\%}} & 77.83\% & \textit{\underline{89.85\%}} & \textbf{66.86\%} & \textit{\underline{81.18\%}} & \textit{\underline{83.97\%}} & \textit{\underline{85.54\%}} & \textit{\underline{81.96\%}} & \textit{\underline{84.03\%}} \\
gemini-3.1-flash-lite-preview & 76.55\% & 77.74\% & 68.87\% & 77.01\% & 53.14\% & 71.43\% & 73.40\% & 70.21\% & 72.18\% & 72.01\% \\
gemma-4-31b-it & 20.00\% & 51.41\% & 35.85\% & 26.57\% & 1.14\% & 30.35\% & 38.89\% & 35.91\% & 44.12\% & 39.33\% \\
gpt-5.4-mini & 68.28\% & 61.76\% & 50.94\% & 65.67\% & 48.57\% & 59.78\% & 60.63\% & 59.70\% & 63.94\% & 61.27\% \\
gpt-5.4-nano & 41.38\% & 42.01\% & 39.15\% & 40.30\% & 26.29\% & 38.62\% & 43.21\% & 35.77\% & 39.20\% & 38.15\% \\
qwen3.5-35b-a3b & 86.21\% & 74.92\% & \textit{\underline{75.00\%}} & 88.06\% & 56.57\% & 77.32\% & 80.14\% & 80.14\% & 77.32\% & 80.14\% \\
qwen3.5-flash & 83.45\% & 77.12\% & 73.11\% & 83.88\% & 55.43\% & 75.89\% & 74.22\% & 76.66\% & 75.78\% & 75.96\% \\
\bottomrule
\end{tabular}%
}
\end{table*}

\begin{table*}[tbp]
\centering
\caption{\textbf{Performance before reasoning under In-Depth evaluation.} The best-performing model in each category is typeset in \textbf{boldface}, and the runner-up is indicated with \textit{underline}.}
\label{tab:appendix_before_depth}
\setlength{\tabcolsep}{3pt}
\resizebox{\textwidth}{!}{%
\begin{tabular}{lcccccccccc}
\toprule
\multirow{2}{*}{\textbf{Model}} & \multicolumn{6}{c}{\textbf{Main}} & \multicolumn{4}{c}{\textbf{Robustness}} \\
\cmidrule(lr){2-7} \cmidrule(lr){8-11}
& \textbf{GEOMETRY} & \textbf{MULTICHART} & \textbf{STATISTICS} & \textbf{TABLE} & \textbf{TIMESEQ} & \textbf{Total} & \textbf{IMAGESTYLE} & \textbf{NOISE} & \textbf{RESOLUTION} & \textbf{Total} \\
\midrule
doubao-seed-2.0-lite & \textit{\underline{66.90\%}} & 65.52\% & \textit{\underline{64.15\%}} & 75.52\% & \textit{\underline{41.71\%}} & \textit{\underline{64.76\%}} & 60.00\% & 77.12\% & \textit{\underline{82.39\%}} & 76.76\% \\
doubao-seed-2.0-pro & \textbf{68.97\%} & \textbf{71.79\%} & \textbf{66.04\%} & \textbf{82.09\%} & \textbf{49.71\%} & \textbf{70.07\%} & \textbf{86.41\%} & \textbf{88.50\%} & \textbf{89.20\%} & \textbf{88.39\%} \\
gemma-4-31b-it & 56.55\% & 54.23\% & 40.09\% & 40.00\% & 17.14\% & 42.50\% & 34.49\% & 31.94\% & 38.50\% & 34.55\% \\
gpt-5.4-mini & 53.79\% & 53.29\% & 51.89\% & 54.63\% & 24.00\% & 49.16\% & 57.84\% & 61.44\% & 62.72\% & 61.27\% \\
gpt-5.4-nano & 40.69\% & 37.30\% & 26.89\% & 40.30\% & 17.14\% & 33.73\% & 39.02\% & 37.75\% & 41.64\% & 39.26\% \\
mimo-v2.5 & 60.00\% & \textit{\underline{66.14\%}} & 59.43\% & \textit{\underline{76.12\%}} & 30.86\% & 61.80\% & 77.35\% & 76.77\% & 75.44\% & 76.42\% \\
qwen3.5-35b-a3b & 62.07\% & 57.05\% & 60.85\% & 75.82\% & 41.14\% & 61.30\% & \textit{\underline{78.75\%}} & \textit{\underline{78.51\%}} & 79.44\% & \textit{\underline{78.86\%}} \\
qwen3.5-flash & 58.62\% & 57.05\% & 52.83\% & \textit{\underline{76.12\%}} & 36.00\% & 58.77\% & 67.94\% & 77.82\% & 79.44\% & 76.71\% \\
\bottomrule
\end{tabular}%
}
\end{table*}